\documentclass[preprint,12pt]{elsarticle}

\makeatletter
\def\ps@pprintTitle{%
	\let\@oddhead\@empty
	\let\@evenhead\@empty
	\let\@oddfoot\@empty
	\let\@evenfoot\@empty
}
\makeatother

\usepackage{amssymb}
\usepackage{amsmath}
\usepackage{lineno}

\usepackage{xspace}
\usepackage{xcolor}
\usepackage{changes}
\usepackage{multirow}
\usepackage{multicol}
\setaddedmarkup{\textcolor{purple}{#1}}

\usepackage{caption}
\newcommand{\Style}[2]{\ensuremath{\textit{#1}_{#2}}{}\xspace}

\newcommand{\EarlySubsc}{va}
\newcommand{\LateSubsc}{tr}
\newcommand{\DomainEarlyStage}{\Style{W}{\EarlySubsc}}
\newcommand{\DomainLateStage}{\Style{W}{\LateSubsc}}
\newcommand{\BackGroundFramesEarlyDomain}{\Style{B}{\EarlySubsc}}
\newcommand{\BackGroundFramesLateDomain}{\Style{B}{\LateSubsc}}
\newcommand{\SimulatedTrain}{\Style{SYN}{tr}}
\newcommand{\SimulatedValid}{\Style{SYN}{va}}
\newcommand{\RealEarlyForegrounds}{\Style{R}{\EarlySubsc}}
\newcommand{\RealLateForegrounds}{\Style{R}{\LateSubsc}}
\newcommand{\FakeEarlyForegrounds}{\Style{F}{\EarlySubsc}}
\newcommand{\FakeLateForegrounds}{\Style{F}{\LateSubsc}}
\newcommand{\RotateAugTrain}{\Style{Rotated}{tr}}
\newcommand{\RotateAugValid}{\Style{Rotated}{va}}
\newcommand{\PseudoSetTrain}{\Style{Pseudo}{tr}}
\newcommand{\PseudoSetValid}{\Style{Pseudo}{va}}
\newcommand{\TestLateDomain}{\Style{LateStage}{te}}
\newcommand{\TestGlobalDomain}{\Style{GHD}{te}}
\newcommand{\PredictionGlobalTest}{\Style{GHD}{qa}}
\newcommand{\PredictionHandHeld}{\Style{HandHeld}{qa}}
\newcommand{\BaseModel}{\text{BaseModel}{}\xspace}
\newcommand{\SYNModel}{\text{SynModel}{}\xspace}
\newcommand{\ROAModel}{\text{RoAModel}{}\xspace}
\newcommand{\PSEModel}{\text{PseModel}{}\xspace}
\newcommand{\CocoModel}{\text{CocoModel}{}\xspace}
\newcommand{\ColorMap}{\text{GLMask}\xspace}
\newcommand{\InstSeg}{instance segmentation\xspace}

\usepackage{array}
\usepackage{multirow}
\usepackage{graphicx}
\newlength{\CellHeight}
\newlength{\CellWidth}
\begin{document}

\begin{frontmatter}

\title{From Semantic To Instance: A Semi-Self-Supervised Learning Approach}

\author[1]{Keyhan Najafian\corref{cor1}}
\author[2]{Farhad Maleki}
\author[1]{Lingling Jin}
\author[1]{Ian Stavness\corref{cor1}}
\affiliation[1]{organization={Department of Computer Science, University of Saskatchewan},
            addressline={176 Thorvaldson Bldg, 110 Science Place}, 
            city={Saskatoon},
            postcode={S7N 5C9}, 
            state={Saskatchewan},
            country={Canada}}
\affiliation[2]{organization={Department of Computer Science, University of Calgary},
            addressline={602 ICT Building, 2500 University Drive NW}, 
            city={Calgary},
            postcode={T2N 1N4}, 
            state={Alberta},
            country={Canada}}
\cortext[cor1]{Keyhan Najafian. Email: keyhan.najafian@usask.ca}
            
%% Abstract
\begin{abstract}
Instance segmentation is essential for applications such as automated monitoring of plant health, growth, and yield. However, extensive effort is required to create large-scale datasets with pixel-level annotations of each object instance for developing \InstSeg models that restrict the use of deep learning in these areas. This challenge is more significant in images with densely packed, self-occluded objects, which are common in agriculture. 
To address this challenge, we propose a semi-self-supervised learning approach that requires minimal manual annotation to develop a high-performing instance segmentation model. We design \ColorMap, an image-mask representation for the model to focus on shape, texture, and pattern while minimizing its dependence on color features. We develop a pipeline to generate semantic segmentation and then transform it into instance-level segmentation. The proposed approach substantially outperforms the conventional instance segmentation models, establishing a state-of-the-art wheat head instance segmentation model with mAP@50 of $98.5\%$. Additionally, we assessed the proposed methodology on the general-purpose Microsoft COCO dataset, achieving significant performance improvement of over $12.6\%$ mAP@50. This highlights that the utility of our proposed approach extends beyond precision agriculture and applies to other domains, specifically those with similar data characteristics.
\end{abstract}

%% Keywords
\begin{keyword}
Instance Segmentation \sep Semi-Supervised Learning \sep Self-Supervised Learning \sep Wheat Head Instance Segmentation \sep Precision Agriculture \sep Data Synthesis
\end{keyword}

\end{frontmatter}

%
% -----------------------------------------
\section{Introduction}\label{sec:intro}
Semantic segmentation~\cite{ronneberger2015u, thisanke2023semantic} is the pixel-level classification task in which each pixel within an image is assigned to a predefined semantic category, such as sky, tree, or vehicle. In contrast, instance segmentation~\cite{he2017mask, Kirillov2023SegmentA} involves identifying and outlining individual objects with well-defined boundaries, even if they belong to the same semantic category. \par 
Instance segmentation poses distinct challenges compared to semantic segmentation, as it demands precise identification and separation of individual objects, which is crucial for applications such as autonomous driving~\cite{Wang2024InstanceSF}, medical imaging~\cite{Deng2023SegmentAM}, and precision agriculture~\cite{Gao2022AutomaticTD}. \par 
Instance segmentation offers a significant advantage in precision agriculture by providing precise identification of individual objects within an agricultural setting~\cite{Charisis2024DeepLI}. In contrast to semantic segmentation, it enables the differentiation of individual plants, fruits, or other agricultural product elements. This distinction is critical for tasks such as tracking and counting crops~\cite{fourati2021wheat, Sapkota2024ComparingYA}, assessing plant health~\cite{Yue2023ImprovedYN, Nayak2024ImprovedDO}, and monitoring growth patterns~\cite{Yue2023ImprovedYN}, resulting in more targeted interventions, better resource allocation, and improved yield predictions. However, training deep \InstSeg models is challenging due to the need for large-scale datasets with pixel-level annotations of each object instance and the significant computational demand involved. \par 
Creating large-scale annotated datasets across various domains is often expensive, time-consuming, and requires rare and highly specialized expertise for data collection and annotation. Moreover, supervised learning introduces challenges such as annotation bias~\cite{chen2021understanding}, lack of generalizability~\cite{Wang2021GeneralizingTU}, and robustness~\cite{hendrycks2021many}. 
Semi-supervised and self-supervised learning leverage partially labeled and unlabeled data, respectively, for model training. They present successful alternatives to supervised learning in addressing its inherent challenges.  By reducing the reliance on manual data annotation, semi- and self-supervised learning approaches are greatly expanding the applicability of deep learning in domains where creating annotated datasets is impractical. \par 
In precision agriculture~\cite{Marinello2023ThePT}, a data-driven approach to optimizing the monitoring and management of agricultural products, the application of semi- and self-supervised learning can improve the performance of deep learning techniques~\cite{najafian2021semi, Chin2023PlantDD, Thangaraj2023ACS,najafian2023semi}. However, the considerable variability in agricultural products and environments, influenced by factors such as weather conditions, lighting, and crop diversity, necessitates the creation of large and diverse annotated databases for effective model training, posing challenges even for semi- and self-supervised learning approaches~\cite{Thangaraj2023ACS}. 

In contrast to general-purpose datasets~\cite{ILSVRC15, Lin2014MicrosoftCC, Kirillov2023SegmentA}, agricultural data are often characterized by small, repetitive, and overlapping objects with hidden and fine boundaries~\cite{najafian2023semi, CossioMontefinale2024CherryCD}, making the manual pixel-accurate and object-specific annotations of such large-scale datasets infeasible. \par 
This paper proposes a semi-self-supervised learning approach for~\InstSeg of images featuring dense, self-similar, and overlapping objects resembling those found in precision agriculture. We experiment and evaluate the effectiveness of our proposed methodology on the wheat head~\InstSeg task. In addition, we experimented with the well-known COCO dataset~\cite{lin2015microsoft} to showcase the applicability of our proposed method in the general-purpose domains. \par 
We first use a limited number of manually annotated images to generate large-scale synthetic datasets with pixel-accurate annotations for each instance. These datasets are then used to develop a \InstSeg model for segmenting images of wheat fields taken in uncontrolled environments. We also apply domain adaptation to address the domain shift between synthetic and real data. Moreover, we propose to use the input RGB image and its corresponding semantic segmentation mask to create an image-mask representation, referred to as~\ColorMap, ultimately enabling the deep model to transform the semantic segmentation mask into an instance mask. 
The~\ColorMap is constructed by combining multiple color spaces derived from the RGB format and a computationally generated segmentation map. More specifically, \ColorMap represents the \textbf{G}rayscale from the RGB channels, \textbf{L} channel of LAB, and a semantic segmentation \textbf{M}ask. 
Models that are overly reliant on color features are problematic in a diverse wheat dataset, where wheat plants change color due to growth stage and variable outdoor lighting. Therefore, the G-L components in ~\ColorMap are designed to encourage the model to focus on shape, texture, and pattern while minimizing its dependence on color features. \par
Specifically, the proposed approach follows a semi-self-supervised learning paradigm, wherein only a small number of pixel-accurate and instance-specific human-annotated images and video frames are used to computationally generate two large-scale datasets. Then a high-performing YOLOv9~\cite{Wang2024YOLOv9LW} model is trained in two consecutive pretraining and domain adaptation phases by leveraging the synthetic datasets. Additionally, we augment the supervisory signals, generated by a pretrained semantic segmentation model within the input data by incorporating~\ColorMap as the model input. Furthermore, we experiment with pseudo-labeling, as a self-supervised approach, utilizing the labels generated by our pretrained model. \par
\begin{figure}[!th]
    \centering
    \includegraphics[width=\textwidth]{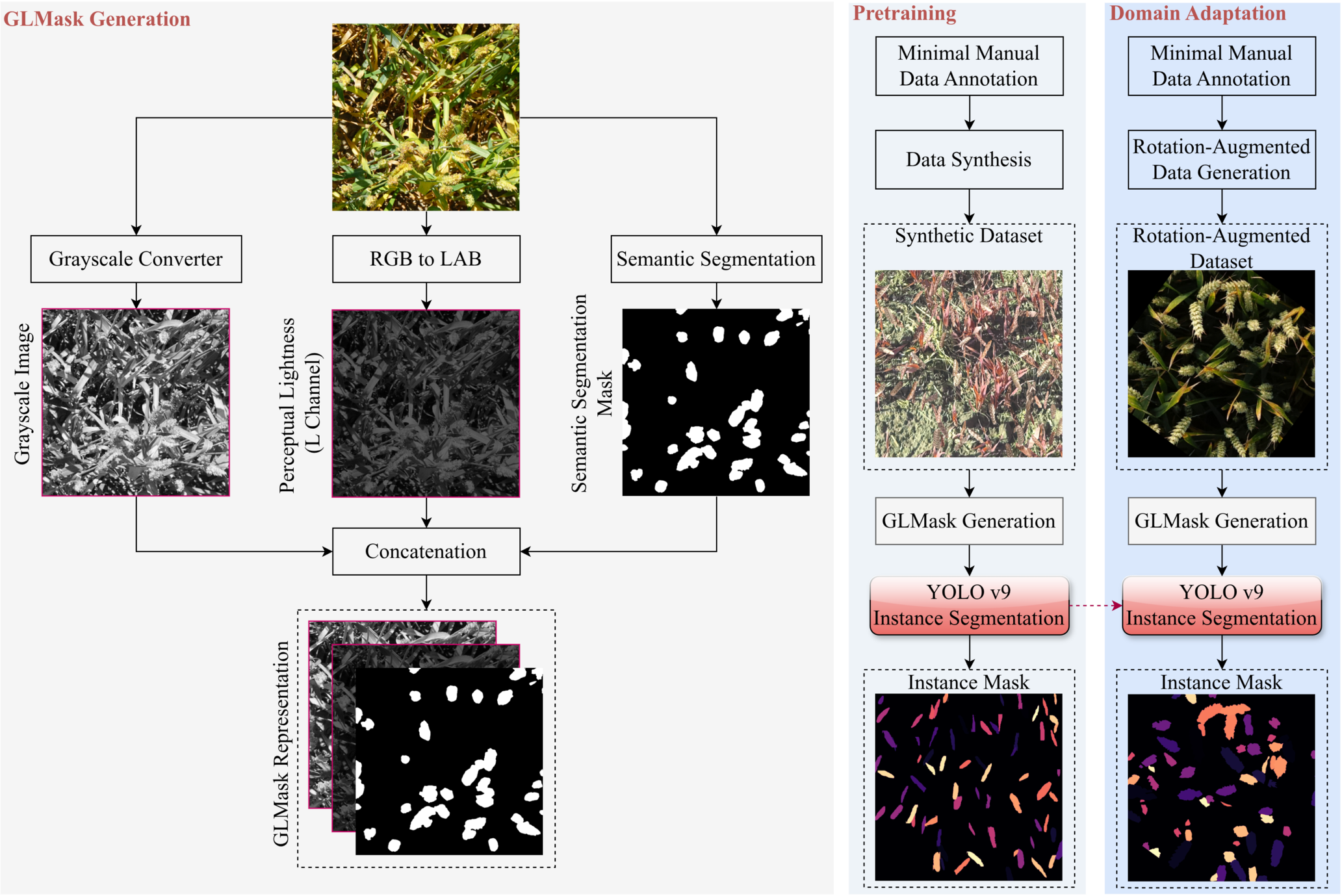}
    \caption{Schematic overview of the proposed Semi-Self-Supervised Learning Framework. The methodology proceeds in three logical stages: (Left) GLMask Representation: To reduce color dependency and enforce structural learning, the input RGB image is decomposed into Grayscale and L-channel (of CIELAB) components, then concatenated with a Semantic Segmentation Mask prior. (Right) Two-Stage Training Strategy: The framework proceeds with (1) Synthetic Pre-training, where a YOLOv9 model is trained on a large-scale synthetic dataset generated via a cut-and-paste pipeline (Sec~\ref{subsec:data_synthesis}); followed by (2) Domain Adaptation, where the model weights are transferred (horizontal dotted arrow) and fine-tuned on a rotation-augmented real dataset to bridge the domain gap (Sec~\ref{sebsec:data_generation}).
    }
    \label{fig:01_overview}
\end{figure}

This work establishes a state-of-the-art top-view wheat head \InstSeg benchmark through the following contributions: (1) introducing a semi-self-supervised learning methodology for high-performance \InstSeg model development; (2) incorporating a shape/pattern emphasized image-mask representation for semantic to instance mask conversion, replacing conventional RGB images, which substantially enhances model performance by providing a strong self-supervisory signal; (3) creating a large-scale computationally annotated synthetic data for deep model training; and (4) a comparison of two domain adaptation methods, rotation augmentation and pseudo-labeling; (5) An evaluation of our proposed methodology on a general-purpose dataset for the instance segmentation task, demonstrating its applicability beyond domain-specific applications. Figure~\ref{fig:01_overview} provides an overview of our proposed semi-self-supervised learning approach. Nevertheless, the utility of our proposed approach extends beyond wheat crops and precision agriculture, making it applicable to any domain involving data with similar characteristics. \par 

%
% -----------------------------------------
\section{Related Works}\label{sec:related_work}
Precision agriculture employs advanced technology to conduct several agricultural analyses through both offline and online computer vision tasks~\cite{Thangaraj2023ACS}. In this domain, image segmentation is used to develop automatic tools~\cite{Nayak2024ImprovedDO, Vyas2023AdvancingPA} and equip machines for monitoring and controlling crops, weeds, soil, as well as crop infection and growth in finer details~\cite{Gupta2023MulticlassWI, Ghanbari2024SemiSelfSupervisedDA, CossioMontefinale2024CherryCD, Myers2024EfficientWH}.\par 
In this context, numerous studies have explored the utility of wheat head semantic segmentation in top-view images of real fields. Rawat et al.~\cite{Rawat2022HowUI} experimented with the wheat head semantic segmentation task with a pool-based active learning approach, obtaining an Intersection over Union (IoU) of $0.70$ on only the 2020 version of the UTokyo subset of the Global Wheat Head Detection (GWHD) dataset~\cite{david2021global}. Najafian et al.~\cite{najafian2023semi} proposed a semi-self-supervised approach for wheat head segmentation, generating synthetic data from limited annotations and employing domain adaptation, resulting in a Dice score of $0.91$. Myers et al.~\cite{Myers2024EfficientWH} enhanced synthetic data realism by training a CycleGAN to transform unrealistic synthetic data into more authentic images, achieving a Dice score of $0.796$ on the GWHD dataset~\cite{david2021global} after training the model on the GAN-generated data and finally applying a pseudo-learning phase. In addition, Ghanbari et al.~\cite{Ghanbari2024SemiSelfSupervisedDA} studied the need for domain adaptation and data diversity, creating large-scale simulated data in training deep models, evaluated specifically on a wheat head segmentation task. Ennadifi et al.~\cite{Ennadifi2022LocalUW} developed a DeepMAC~\cite{Ennadifi2022LocalUW} and YOLOv5~\cite{Wang2022YOLOv7TB} model for single-head segmentation of wheat heads, individually extracted using wheat head bounding boxes from $701$ manually annotated images or YOLOv5 model-generated bounding boxes, achieved an average F1 score of $86\%$ with their best-performing model. \par 
Instance Segmentation, a finer approach than semantic segmentation, goes beyond identifying regions and delineates individual object instances within each category~\cite{Marks2023HighPL, Gao2022AutomaticTD}. Particularly, recent advancements in foundation models for pixel-level classification tasks have greatly increased the use of \InstSeg across various domains, including precision agriculture, through fine-tuning or zero-shot learning. Particularly, given the challenges associated with data annotation, supervised fine-tuning of these pretrained models offers a more straightforward approach, which requires less data annotation, compared to training a model from scratch, which demands a substantial number of annotated samples. Furthermore, the availability of manually annotated datasets, such as GWHD~\cite{david2021global}, serves as another crucial component in optimizing pretrained deep learning models for the agricultural domain tasks. \par
Li et al.~\cite{Li2023EnhancingAI} utilized transfer learning to fine-tune the SAM model~\cite{Kirillov2023SegmentA} for crop disease and pest segmentation in agricultural domains. They employed three datasets: two for coffee-leaf diseases, incorporating $1,100$ images, and one for pest image segmentation, containing $5,464$ images of $10$ common agricultural pests. The coffee-leaf disease datasets were split into training and testing subsets with a $80/20$ ratio. The pest dataset was divided into two groups, each with five pest categories. Their final model achieved an IoU of $59.58\%$ on the coffee-leaf disease datasets and $89.56\%$ on the pest datasets. \par
Nayak et al.~\cite{Nayak2024ImprovedDO} fine-tuned a YOLOACT~\cite{Bolya2019YOLACTRI} \InstSeg model to locate and segment wheat spike regions infected by Fusarium Head Blight (FHB). They employed a transfer learning approach, utilizing a dataset comprising $1,800$ annotated side-view images, which included $600$ images of infected wheat fields. This work focused exclusively on segmenting small, infected spike regions within the side-view images without segmenting all spikes. \par 
Gao et al.~\cite{Gao2022AutomaticTD} developed a DL-based model for severity assessment of wheat FHB in four healthy, mild, moderate, and severe categories. To do so, two BlendMask models were trained on $2,832$ training and $922$ validation sub-image samples for FHB-diseased single-spike segmentation and $524$ training and $166$ validation whole spike segmentation of side-view images. The samples were captured in sunny, cloudy, rainy, etc. environmental conditions and with backgrounds such as soil, other spikes, sky, and clouds. However, in the whole wheat field image segmentation, only sharp images in the foreground were manually annotated and, thereby, segmented by the model. The final models achieve $48.23\%$ IoU for spike \InstSeg and a $52.41\%$ IoU for FHB disease detection. The severity assessment was conducted by comparing the segmentation map of the infected regions with the segmentation map of the entire spike. \par
Despite the similarities to the aforementioned studies, our work stands apart in several key aspects. We focus on the \InstSeg of entire wheat head objects within large-scale, high-resolution, and top-view wheat-field images characterized by dense, self-similar, and overlapping objects. In contrast to most current research that relies on supervised learning methodologies and extensive annotations, our approach leverages semi-self-supervised learning and minimal data annotation. However, similar to other works, we utilize pretrained models~\cite{Kirillov2023SegmentA,Wang2024YOLOv9LW} to facilitate the development process, avoiding the need to train such large models from scratch, and leveraging the transfer of their learned knowledge. \par
With regard to the advancements in state-of-the-art (SOTA) \InstSeg models, such as the Segment Anything Model (SAM) series~\cite{Kirillov2023SegmentA, Ravi2024SAM2S}, we opted to utilize YOLOv9~\cite{Wang2024YOLOv9LW}---recognized as the SOTA in object detection and localization, known for its accuracy and efficiency in instance segmentation---in this work. SAM, which has outperformed models including FastSAM~\cite{Zhao2023FastSA}, was extensively trained on over $1.1$ billion segmentation masks across 11 million licensed images. FastSAM, on the other hand, leverages YOLOv8~\cite{reis2023real} and a CLIP~\cite{Ravi2024SAM2S} prompting head, achieving $50\times$ better speed but with reduced performance compared to SAM. SAM2~\cite{Ravi2024SAM2S} introduced memory mechanisms for real-time image and video segmentation. However, YOLOv9 stands out by introducing the integration of Programmable Gradient Information (PGI) to address the information bottleneck problem and proposing the Generalized Efficient Layer Aggregation Network (GELAN) for optimal parameter utilization and computational efficiency. This architecture's capacity to flexibly integrate various computational blocks while maintaining high performance makes YOLOv9 particularly adaptable to a wide range of applications, including \InstSeg, aligning better with our work's objectives. \par

% -----------------------------------------
\section{Material and Methods}\label{sec:method}
We propose a Semi-Self-Supervised learning framework designed to segment densely packed objects with minimal manual annotation. As illustrated in Figure~\ref{fig:01_overview}, our approach operates in three stages: (1) Representation Learning, where we introduce \textit{GLMask} to shift the model's focus from color to structure; (2) Synthetic Pre-training, utilizing a cut-and-paste synthesis pipeline to initialize weights; and (3) Domain Adaptation, employing a rotation-based consistency strategy to bridge the gap between synthetic and real-world data. This pipeline enables the transition from coarse semantic signals to precise instance masks. To operationalize this framework, specifically the data synthesis and generation phases, a curated set of minimal real-world samples is required. We therefore begin by detailing these fundamental data requirements.

\subsection{Data}\label{subsec:data}
\begin{figure}[!th]
    \centering
    \includegraphics[width=\textwidth]{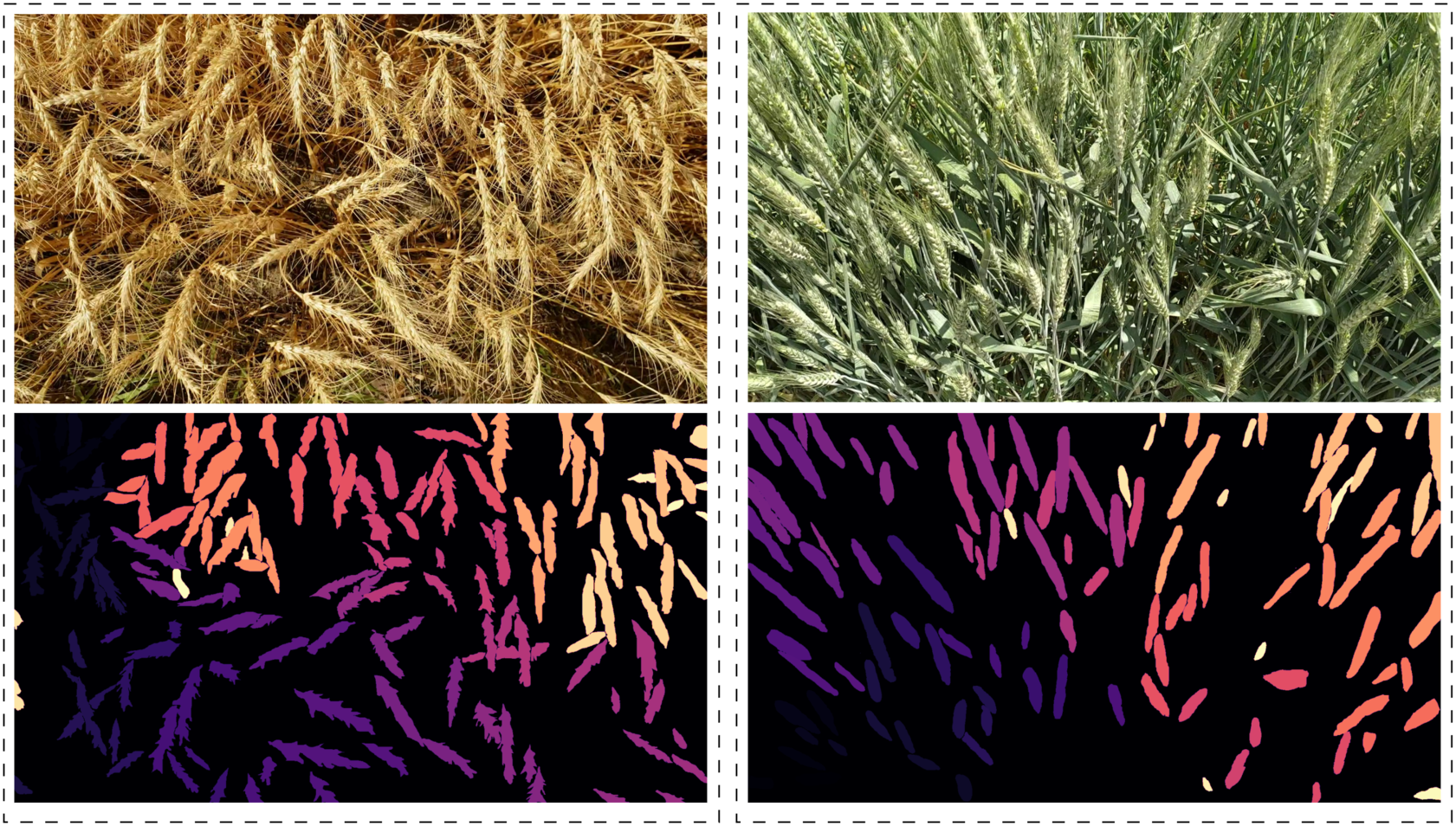}
    \caption{Examples of video frames from the \textit{Harvest-ready} growth stage (\DomainLateStage, left) and \textit{Heading-complete} stage (\DomainEarlyStage, right) wheat field domains, along with their corresponding human-annotated instance-specific masks (bottom row).
    }
    \label{fig:02_synthesis_frames}
\end{figure}

We manually annotated 10 image frames, each sized $1080 \times 1920$, taken from two top-view video clips of wheat fields from Ghanbari et al.~\cite{Ghanbari2024SemiSelfSupervisedDA} for data synthesis (section~\ref{subsec:data_synthesis}). According to ~\cite{miller1992growth}, these images represent two domains corresponding to different growth stages of wheat. Eight images, denoted as~\DomainLateStage, were from a Wheat Mature and \textit{Harvest-ready} growth stage, while the remaining two images, denoted as \DomainEarlyStage, were from a \textit{Heading-complete} growth stage. Figure~\ref{fig:02_synthesis_frames} illustrates sample images from these two domains. Images in \DomainLateStage and \DomainEarlyStage were used for synthesizing a training set and a validation set, respectively. Each frame contained $106.7$ wheat head instances on average with the number ranging from $14$ to $262$. \par
Additionally, we utilized $29$ background video clips, without any wheat plant, of which we used $15$ videos to extract a set of $50,072$ frames, referred to as \BackGroundFramesLateDomain, for the synthesizing training set. From the remaining $14$ video clip, we extracted $13,253$ image frames (referred to as \BackGroundFramesEarlyDomain) for the validation set data synthesis. Images in \BackGroundFramesLateDomain and \BackGroundFramesEarlyDomain were randomly cropped to a size of  $1024 \times 1024$. \par 
In addition, we used $19$ semantically annotated pairs of real field images from Najafian et al.~\cite{najafian2023semi} and manually converted their annotation to instance masks. From these, $18$ pairs of these image/mask samples were from the GWHD~\cite{david2021global} dataset, and the remaining pair was extracted from \DomainLateStage. 
The 19 pairs were also divided into two subsets, referred to as \RotateAugTrain and \RotateAugValid, each containing one sample from every pair. We then used~\RotateAugTrain and~\RotateAugValid to generate two rotation-augmented datasets, as detailed in section~\ref{sebsec:data_generation}. Each $1024 \times 1024$ image in$~\RotateAugTrain \cup \RotateAugValid$ contains an average of $52.8$ instances, with instance counts varying between $16$ and $91$.\par

\begin{figure}[!th]
    \centering
    \includegraphics[width=\textwidth]{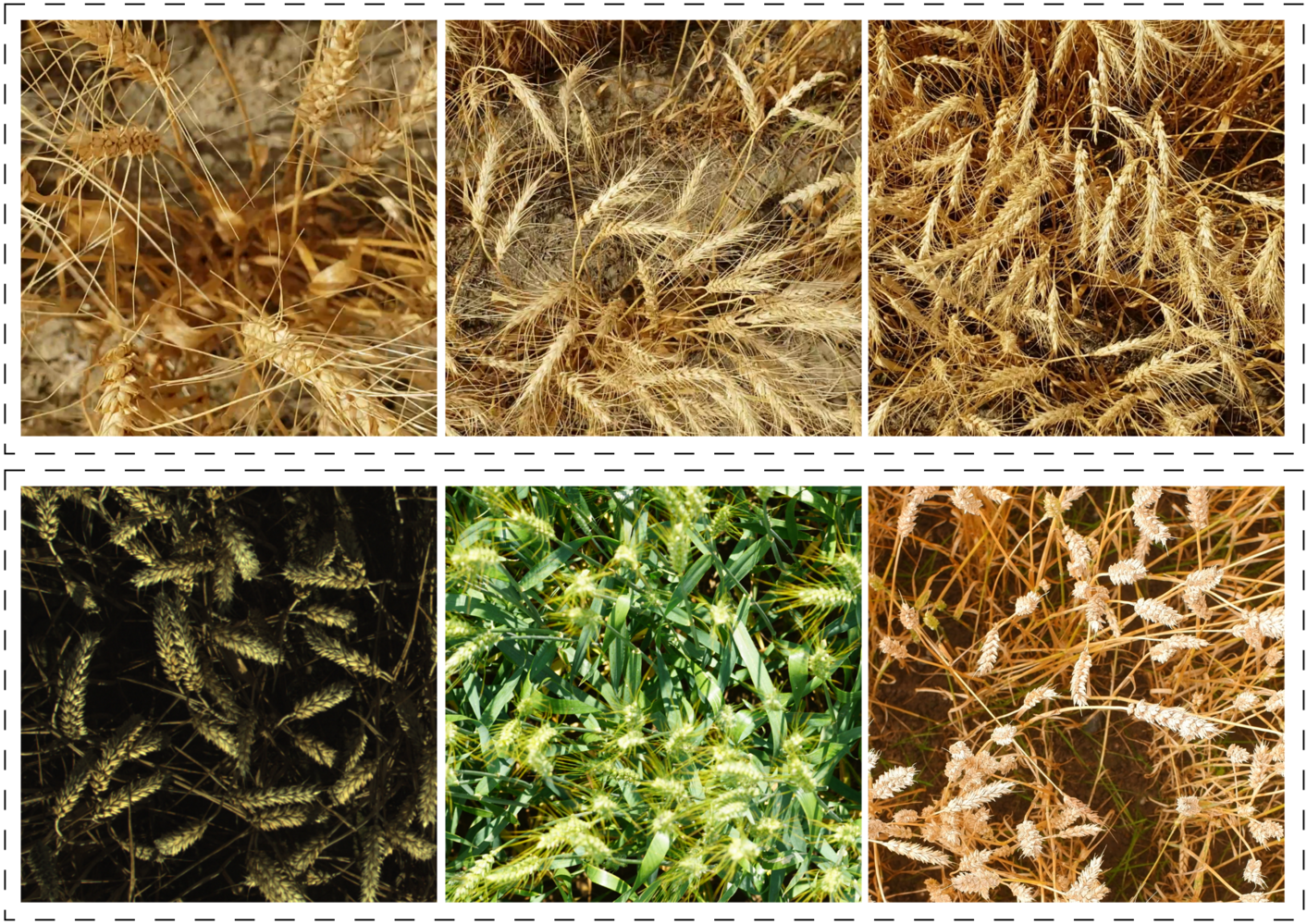}
    \caption{
        Examples illustrating the diversity within our human-annotated test sets. The top row represents samples from our single-domain test set~\TestLateDomain. The bottom row shows samples from our $18$ domains test set~\TestGlobalDomain.
    }
    \label{fig:03_test_samples}
\end{figure}

For model evaluation, we manually annotated a set of $365$ images, hereafter referred to as \TestGlobalDomain. The average number of wheat head instances for these images was $52.3$, ranging from $0$ to $102$.
Also, a set of 100 images of size $1024 \times 1024$, referred to as \TestLateDomain, were used for further model evaluation. Each frame in~\TestLateDomain has an average of $58.12$ instances, ranging from $5$ to $146$. Figure~\ref{fig:03_test_samples} presents a few manually annotated examples from our evaluation sets.\par
We utilized the unannotated images from the GWHD dataset~\cite{david2021global}, which lacked instance annotations. These images have been partitioned into three subsets of \PseudoSetTrain, \PseudoSetValid, and \PredictionGlobalTest, with $3657$, $1476$, and $1382$ images, respectively~\cite{david2021global}. We excluded the 365 images in \TestGlobalDomain from \PseudoSetTrain.
We selected two external unannotated datasets for qualitative assessment: (1) $180$ samples, comprising $10$ samples from each of the $18$ domains from \PredictionGlobalTest and (2) $100$ video frames, with $10$ image frames extracted from $10$ different videos representing distinct wheat field domains, denoted as~\PredictionHandHeld. These videos were captured with high-resolution handheld cameras and were center-cropped to $1024 \times 1024$ pixels.\par
%
%
% ----------------------------
\subsection{\ColorMap Representation}\label{subsec:glm}
As the foundational component of our pipeline, we design an input format that explicitly leverages semantic priors to guide the transformation from semantic to instance-level segmentation. 
Specifically, we utilized the Grayscale (G) and the perceptual lightness channel of LAB (L) color space versions of each RGB image, along with the corresponding semantic mask (Mask), to create a new image-mask representation of the RGB image, referred to as~\ColorMap. 

For the semantic mask component, we used pseudo-masks generated by the model developed for wheat head semantic segmentation by Najafian et al.~\cite{najafian2023semi}, which reportedly achieves a Dice score of 0.91 on wheat datasets. By embedding the semantic mask, we aimed at efficient segmentation transfer from semantic to instance masks. Additionally, we designed~\ColorMap to address a common issue in convolutional neural networks, where models often become overly reliant on color features. This reliance can hinder model generalization in the context of a diverse wheat dataset, as wheat color varies significantly due to growth stages and changing outdoor lighting conditions. By incorporating grayscale and the L channel of the LAB color spaces,~\ColorMap intentionally reduces color information, encouraging the model to prioritize shape, texture, and pattern features. \par
The grayscale image $G$, derived by weighting the RGB images using $G = 0.2125 \times R + 0.7154 \times G + 0.0721 \times B$ formula~\cite{Poynton1997FrequentlyAQ}, represents brightness, which is a subjective visual perception influenced by both luminance and color. It enhances contrast and improves edge and shape visibility. $G$ also reduces color-induced noise, thereby improving the quality of agricultural images, which generally contain higher noise levels than general-view images. \par 
On the other hand, the L channel of the LAB color space represents an objective, measurable quantity of perceptual lightness. In contrast to grayscale, it approximates human vision perceptually more accurately in a uniform manner. The L channel separates luminance from chromatic information, improving shadow identification and highlighting shape and structural details. It provides a more balanced representation of light and dark areas while reducing the color-related artifacts.
To create the three-channel \ColorMap image (please see Figure~\ref{fig:01_overview}), we concatenated G, L, and M, where M contains values of 0 or 255. \ColorMap is used as the input for our model in all the training or evaluation phases.  \par 

The GLMask representation was specifically designed as a three-channel input (G, L, M) to address the input requirements of the chosen pre-trained YOLOv9e-Seg architecture while ensuring the model learns features based on shape and density rather than fluctuating spectral signatures. This configuration prioritizes domain-invariant structural information over raw color, which often varies across different growth stages and environments. Consequently, this design choice targets the most informative, color-free components as a principled design constraint, focusing the model's capacity on features that persist across the diverse wheat domains. \par
%
%
%
% ----------------------------
\subsection{Data Synthesis}\label{subsec:data_synthesis}
To effectively train our model using the proposed~\ColorMap representation without relying on extensive manual annotation, we leveraged a synthetic data generation strategy. We adopted the cut-and-paste methodology originally proposed by~\cite{najafian2021semi} for object detection and applied it to the \InstSeg data synthesis with some modifications. As the primary adjustment, we increased the number of manually annotated frames---containing larger wheat head objects---to $10$, categorized as~\DomainLateStage and~\DomainEarlyStage. Additionally, we incorporated a wider range of video frames from various background scenes, sourced from~\BackGroundFramesLateDomain and~\BackGroundFramesEarlyDomain, to enhance the final synthesized data variability. \par
\begin{figure}[!th]
    \centering
    \includegraphics[width=\textwidth]{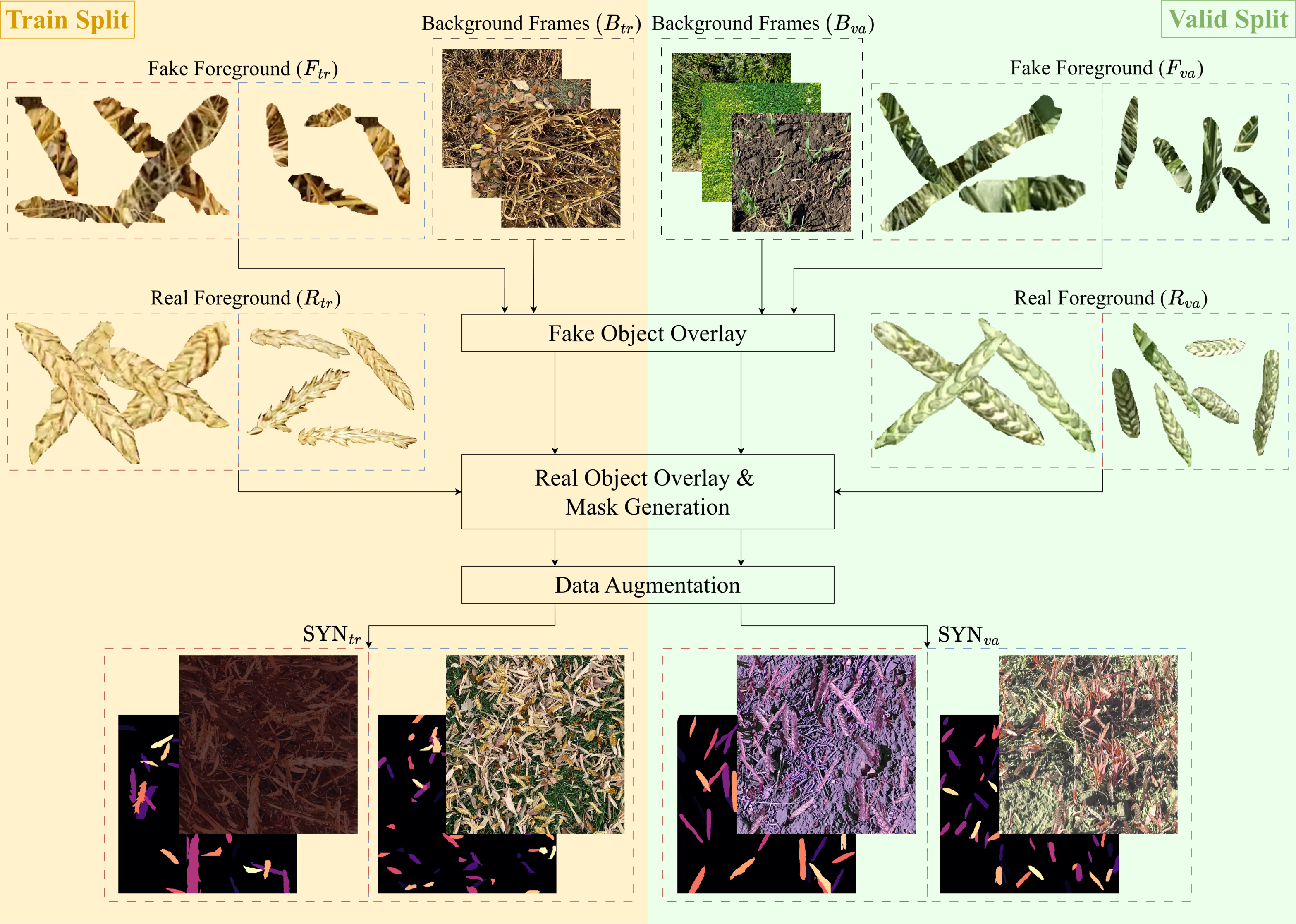}
    \caption{Detailed workflow of the data synthesis pipeline for \InstSeg, adapted from~\cite{najafian2023semi}. The process is strictly partitioned into Train Split (left, yellow) and Valid Split (right, green) streams to ensure distinct data distributions, utilizing wheat heads categorized into \textit{Harvest-ready} (\DomainLateStage) and \textit{Heading-complete} (\DomainEarlyStage) stages along with their corresponding diversified background frames (\BackGroundFramesLateDomain and \BackGroundFramesEarlyDomain). The pipeline integrates three source components: background frames (\ensuremath{B}), synthetic fake foregrounds (\ensuremath{F}), and extracted real foregrounds (\ensuremath{R}). The synthesis proceeds sequentially: (1) Fake Object Overlay: Backgrounds are first populated with synthetic wheat templates to create a dense underlying texture; (2) Real Object Overlay \& Mask Generation: Real wheat instances are superimposed using density-dependent selection logic (using smaller heads in blue dashed boxes for $\ge 50$ objects and larger heads in red dashed boxes for fewer objects) to simulate varying altitudes, while simultaneously generating pixel-perfect instance masks; and (3) Data Augmentation: The composited images undergo photometric and geometric transformations to produce the final synthetic training (\SimulatedTrain) and validation (\SimulatedValid) datasets.
    % Data synthesis pipeline for \InstSeg, adapted from~\cite{najafian2023semi}. To prevent leakage, Train (yellow) and Valid (green) streams use distinct growth stages (\DomainLateStage vs. \DomainEarlyStage) and backgrounds. Inputs---backgrounds (B), fake (F), and real (R) foregrounds---are processed sequentially: (1) Fake Overlay creates dense underlying texture; (2) Real Overlay superimposes instances using density logic (selecting smaller heads in blue boxes for ≥50 objects; larger in red otherwise) to simulate altitude and generate masks; and (3) Augmentation yields the final \SimulatedTrain and \SimulatedValid datasets.
    }
    \label{fig:04_data_synthesis_pipeline}
\end{figure}

In line with the methodology described by~\cite{najafian2021semi, najafian2023semi}, we extracted two sets of fake (\FakeLateForegrounds and~\FakeEarlyForegrounds) and real wheat head objects (\RealLateForegrounds and \RealEarlyForegrounds), which were randomly overlaid on the background images in order. We used~\BackGroundFramesLateDomain,~\FakeLateForegrounds and~\RealLateForegrounds for large-scale training data synthesis of $20,000$ sample size, hereby denoted as~\SimulatedTrain. Additionally, we employed~\BackGroundFramesEarlyDomain,~\FakeEarlyForegrounds and~\RealEarlyForegrounds in synthesizing a validation set of size $10,000$, entitled~\SimulatedValid, with a different distribution from the training set. The data synthesis pipeline is visually demonstrated in Figure~\ref{fig:04_data_synthesis_pipeline}. \par  
To synthesize each sample, we randomly selected $10$ to $100$ fake and then real wheat heads to overlay on the backgrounds. However, dissimilar to~\cite{najafian2023semi}, we divided the real wheat heads into two subsets based on the object's larger dimensions, with smaller and larger wheat heads in a $50\%-50\%$ ratio. Overlaying $10$ to $50$ objects, we selected from the subset containing larger heads; for more than $50$ overlying objects, we chose from the subset with smaller wheat heads. This approach aimed to enhance the realism of the synthesized samples by simulating wheat field images captured from varying altitudes. Additionally, we designed our pipeline to computationally generate the \InstSeg maps and contour labels required for YOLO models, in parallel with synthesizing each image, for model development. \par
For training with the YOLOv9~\cite{Wang2024YOLOv9LW} model, we chose not to apply additional online augmentations, relying solely on the augmentations integrated into YOLOv9. Instead, we increased the variability of our dataset by applying strong augmentations to the synthesized samples, in an offline manner. Specifically, we applied Flip, Rotation, and Elastic transformations for fake and real wheat head objects before overlaying them. Furthermore, we opted for a long list of pixel-level augmentations, including Color Jitter, Channel Shuffle and Dropout, Solarize, Blur, and Noise for augmenting the synthesized images at the end of the process. We employed the Albumentations~\cite{AlbumentationBuslaev} Python package, version \text{1.4.14}, in implementing all the augmentation transformations.
%
%
%
% ----------------------------
\subsection{Domain Adaptation via Rotation-Augmented Data}\label{sebsec:data_generation}
To bridge the inevitable domain shift between the synthetic data generated in the previous step and real-world agricultural environments, we prepared a real-world dataset for domain adaptation.
For this second phase of the model training, we utilized real wheat field images to implement our domain adaptation method. As previously discussed, pixel-accurate instance annotation of many agricultural images is infeasible, and the $19$ samples in~\RotateAugTrain were insufficient for fine-tuning a large model such as YOLOv9. Addressing these limitations, we extended the~\RotateAugTrain dataset by applying image rotation.
To enhance variability in our limited training dataset and better represent natural environments, we utilized rotation to simulate various angled top-down views of bending wheat stems, particularly effective for aerial imaging in windy conditions.
Following the method used by~\cite{najafian2023semi}, each image/mask pair was rotated with $0$ to $259$ degrees, resulting in a large-scale dataset of $6840$ samples. To maintain the authenticity of samples for addressing domain shifts between synthetic and real data, we chose to exclude additional pixel-level or spatial-level transformations, apart from those applied by the YOLOv9 models in an online manner. We repeated the same process for the~\RotateAugValid dataset, expanding it to match the sample size of~\RotateAugTrain. \par 
\begin{table}[!th]
\centering
\footnotesize
\renewcommand{\arraystretch}{1.2}
\caption{
    Summary of the datasets utilized, manually annotated, or computationally generated.
}
\label{tab:dataset_summary}
\begin{tabular}{ccc}
\hline
\textbf{Dataset}                    & \textbf{Aim} & \textbf{Notation}                    \\ \hline
\multirow{2}{*}{Synthetic}          & Train                      & \SimulatedTrain       \\
                                    & Valid                      & \SimulatedValid       \\ \hline
\multirow{2}{*}{Rotation-Augmented} & Train                      & \RotateAugTrain       \\
                                    & Valid                      & \RotateAugValid       \\ \hline
\multirow{2}{*}{Pseudo-Labeled}     & Train                      & \PseudoSetTrain       \\
                                    & Valid                      & \PseudoSetValid       \\ \hline
\multirow{2}{*}{Manually Annotated} & Test                       & \TestLateDomain       \\
                                    & Test                       & \TestGlobalDomain     \\ \hline
\multirow{2}{*}{Unannotated}        & Quality Assessment     & \PredictionHandHeld   \\
                                    & Quality Assessment     & \PredictionGlobalTest \\ \hline
\end{tabular}
% }
\end{table}

%
% ----------------------------
\subsection{Domain Adaptation via Pseudo-Labeling}\label{subsec:pseudo}
To evaluate the efficacy of our primary rotation-based adaptation strategy, we investigated Pseudo-Labeling as a comparative self-supervised baseline.
We explored this widely used approach as an alternative for domain adaptation, as it is known to enhance model generalization by leveraging larger pseudo-labeled datasets, improving performance and reducing overfitting while lowering labeling costs, particularly in domains such as precision agriculture with packed objects.
We utilized the best-performing model in the first phase, trained on the synthetic data, to label the unannotated~\PseudoSetTrain and ~\PseudoSetValid sets. Subsequently, we used these sets to fine-tune the model, initially trained on only synthetic data. We compare the performance of this model with the model developed using \RotateAugTrain and \RotateAugValid datasets to obtain our best-performing model as a better domain adaptation technique. \par 
%
% ----------------------------
\subsection{Experiments}
As described in section~\ref{sec:related_work}, we selected the pretrained extended model YOLOv9e-Seg architecture with $60,512,800$ parameters for all our experiments. We trained and evaluated the model using the data outlined in Table~\ref{tab:dataset_summary}, with training conducted for $100$ epochs in each training phase on $1024 \times 1024$ images. Using the AdamW optimizer~\cite{Loshchilov2017DecoupledWD}, we set an initial learning rate of $1e-3$ to train the model on the~\SimulatedTrain and~\SimulatedValid datasets, resulting in~\SYNModel, and a learning rate of $1e-4$ to fine-tune the~\RotateAugTrain and~\RotateAugValid datasets, leading to the development of~\ROAModel. We also applied a learning rate of $1e-4$ and used the \PseudoSetTrain and \PseudoSetValid datasets to conduct the pseudo-labeling experiment by fine-tuning model~\SYNModel in comparison to model \ROAModel, yielding \PSEModel. We further developed a baseline model,~\BaseModel, identical to~\SYNModel but on RGB images. The batch size was set to $18$ and the IoU to $0.7$. The deterministic and seed parameters were set to ensure reproducibility, and the remaining parameters were kept at their default values. The default data augmentations applied included HSV color augmentation, horizontal flip, mosaic augmentation, erasing, translation, and scaling. \par

We subsequently trained two models on the training split of the Microsoft COCO 2017 dataset, with the same exact parameters as those utilized in the wheat experiments with some exceptions. We utilized the validation split (5k images) exclusively for evaluation to ensure zero data leakage during training. These models were trained on both the RGB and \ColorMap versions of the dataset, with an adjusted batch size of $48$, enabled by a training image size of $640 \times 640$ (up from the batch size of $18$ for $1024 \times 1024$ images in wheat experiments). Given the dataset's substantial size, both COCO models were trained for $50$ epochs. Additionally, we set IoU and conf parameters to be 0.6 and 0.25, respectively, during the evaluation. These two parameters enhanced both RGB and \ColorMap model performances on the COCO validation set. We evaluate the models on the validation set of this dataset, as annotations for its test subset were unavailable. 

Note that we only require the binary (foreground vs. background) mask to create the \ColorMap version of the data. These masks were generated by aggregating all ground-truth instance annotations into a single binary foreground channel. This implies that, unlike the binary mask used as the semantic mask for the single-class wheat dataset, it serves as a binary mask instead of the semantic mask for the COCO dataset, which contains 80 distinct object classes. \par 

All experiments were executed on three GPUs, specifically two \textit{A40 GPUs} each equipped with $48$ GB GDDR6, and one \text{Tesla V100S PCIe} with $32$GB GPU. The environment was configured with Python \text{3.12}, Ultralytics version \text{8.2.79}, and PyTorch \text{2.4.0+cu124}. \par
%
%
% -----------------------------------------
\section{Results}\label{sec:results}
\begin{table*}[!th]
\centering
\footnotesize
\renewcommand{\arraystretch}{1.3}
\caption{
Performance evaluation of \SYNModel, \ROAModel, and \PSEModel on our manually annotated test sets, \TestLateDomain and \TestGlobalDomain. We utilized YOLOv9e-Seg~\cite{Wang2024YOLOv9LW}, denoted as YOLOv9e, as the base pretrained model for our experiments. The Color column indicates whether the experiments were conducted on the RGB version of the data or using our approach with \ColorMap data.
}
\label{tab:models_performances}
\resizebox{\textwidth}{!}{%
    \begin{tabular}{cccccccc}
    \hline
    \textbf{Model}                          & \textbf{Pretrained Model}  & \textbf{Test Data}  & \textbf{Color Space}        & \textbf{P}~$\uparrow$ & \textbf{R}~$\uparrow$ & \textbf{mAP@50}~$\uparrow$ & \textbf{mAP@50-95}~$\uparrow$ \\ \hline
    \multirow{2}{*}{\BaseModel} & \multirow{2}{*}{YOLOv9e}            & \TestLateDomain  & RGB & 81.9       & 38.2       & 49.7            & 24.9               \\
                                            &                                         & \TestGlobalDomain & RGB & 66.9      & 41.9       & 50.4           & 27.1               \\ \hline 
    \multirow{2}{*}{\SYNModel} & \multirow{2}{*}{YOLOv9e}            & \TestLateDomain & \ColorMap  & 86.5       & 82.0       & 88.2            & 61.2               \\
                                            &                                        & \TestGlobalDomain & \ColorMap & 95.2       & 95.3       & 97.9            & 77.0               \\ \hline
    \multirow{2}{*}{\ROAModel} & \multirow{2}{*}{\SYNModel} & \TestLateDomain  & \ColorMap & \textbf{93.2}       & \textbf{93.3}       & \textbf{95.7}            & \textbf{80.7}               \\
                                            &                                          & \TestGlobalDomain & \ColorMap & \textbf{98.1}       & \textbf{96.5}        & \textbf{98.5}            & \textbf{85.3}               \\ \hline
    \multirow{2}{*}{\PSEModel} & \multirow{2}{*}{\SYNModel} & \TestLateDomain & \ColorMap  & 77.1       & 71.7       & 77.5            & 49.0               \\
                                            &                                         & \TestGlobalDomain & \ColorMap & 95.8       & 95.2     & 98.0            & 76.6               \\ \hline 
    \end{tabular}
}
\end{table*}
Table~\ref{tab:models_performances} presents the performances of the developed models using Precision, Recall, and mAP metrics on our test sets. The baseline model~\BaseModel that was trained on the RGB version of the synthetic data (\SimulatedTrain and \SimulatedValid) performed significantly lower than the~\SYNModel, which was trained under the same condition but on the~\ColorMap version of the synthetic data. \par 
\begin{figure}[!th]
    \centering
    \includegraphics[width=\textwidth]{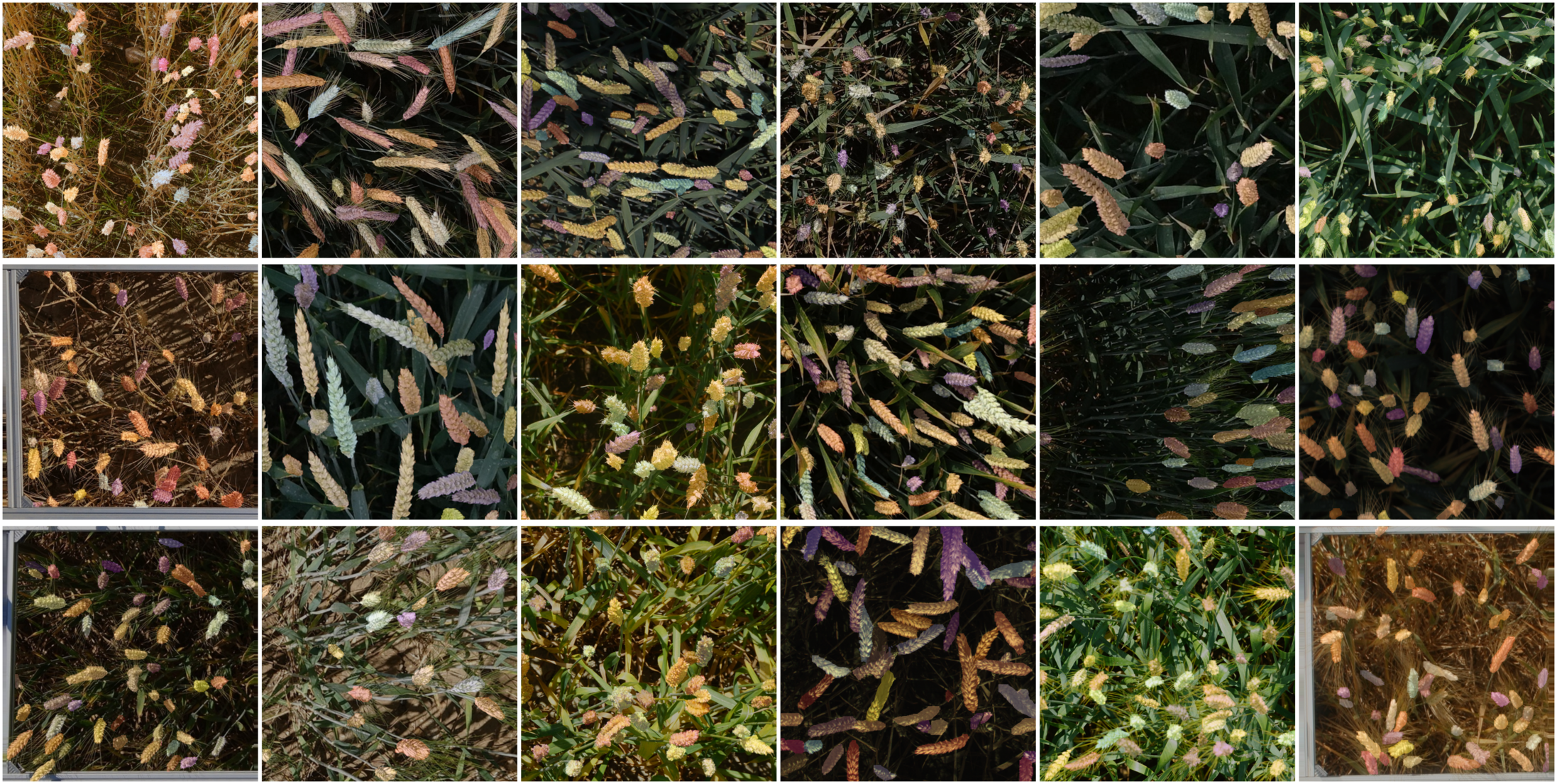}
    \caption{Prediction Performance of \ROAModel across the $18$ domains of \TestGlobalDomain test set.}
    \label{fig:05_quality_asses_our_testset}
\end{figure}
\SYNModel, trained solely on synthetic data (\SimulatedTrain and~\SimulatedValid), obtained mAP@50 of an $88.2\%$, yielding $38.5\%$ improvement over~\BaseModel, on the~\TestLateDomain, representing a single data domain. Despite being trained only on synthetic data, the model achieved $47.5\%$ higher performance in comparison to~\BaseModel with $97.9$ mAP@50 on the~\TestGlobalDomain, which is considered an external test set for this model. Nevertheless, the performance of both~\BaseModel and~\SYNModel on the~\TestGlobalDomain, which serves as an external evaluation set for both models, demonstrate the efficiency of our data synthesis approach for~\InstSeg. \par
\ROAModel was built upon~\SYNModel by being fine-tuned on the rotation-augmented dataset~\RotateAugTrain and \RotateAugValid. \ROAModel obtained the high-performance of $95.7\%$ \text{mAP@50} on our single-domain~\TestLateDomain set and a \text{mAP@50} of $98.5\%$ on \TestGlobalDomain. Figure~\ref{fig:05_quality_asses_our_testset} displays the prediction performance of this model on samples from all $18$ domains of~\TestGlobalDomain. \par

\begin{table}[!hp]
	\caption{Performance evaluation of our \BaseModel (base) and \ROAModel (best) models across the $18$ diverse domains of the $GHD_{te}$ dataset.}
	\label{tab:PerDomainPerformance}
	\scriptsize
	\renewcommand{\arraystretch}{2.8} % Adjusted for better vertical fit
	
	% --- LEFT COLUMN: Domains 00-08 ---
	\begin{minipage}[b]{0.48\textwidth} 
		\centering
		\begin{tabular}{c c c c}
			Domain & Model & Pretrained & mAP@50 \\ \hline\hline
			
			\multirow{2}{*}{\includegraphics[width=\CellWidth, height=\CellHeight]{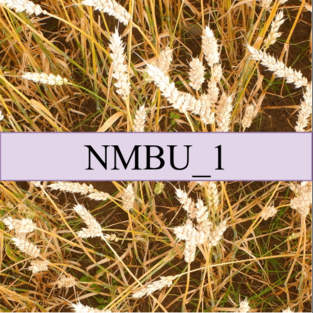}}
			& \BaseModel & YOLOv9e    & 50.5 \\
			& \ROAModel  & \SYNModel & 96.9 \\ \hline
			
			\multirow{2}{*}{\includegraphics[width=\CellWidth, height=\CellHeight]{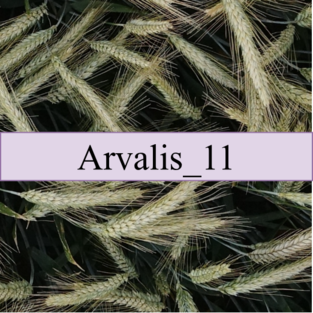}}
			& \BaseModel & YOLOv9e   & 57.5 \\
			& \ROAModel  & \SYNModel & 97.2 \\ \hline
			
			\multirow{2}{*}{\includegraphics[width=\CellWidth, height=\CellHeight]{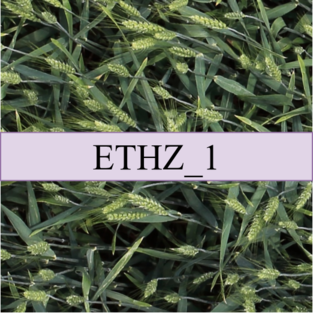}}
			& \BaseModel & YOLOv9e   & 71.1 \\
			& \ROAModel  & \SYNModel & 98.0 \\ \hline
			
			\multirow{2}{*}{\includegraphics[width=\CellWidth, height=\CellHeight]{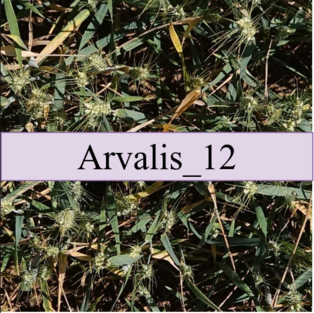}}
			& \BaseModel & YOLOv9e   & 14.9 \\
			& \ROAModel  & \SYNModel & 99.5 \\ \hline
			
			\multirow{2}{*}{\includegraphics[width=\CellWidth, height=\CellHeight]{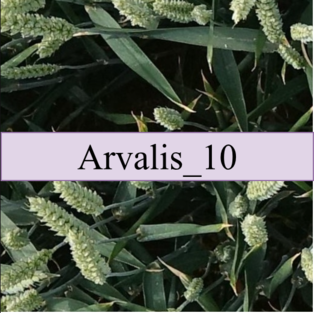}}
			& \BaseModel & YOLOv9e   & 28.9 \\
			& \ROAModel  & \SYNModel & 99.0 \\ \hline
			
			\multirow{2}{*}{\includegraphics[width=\CellWidth, height=\CellHeight]{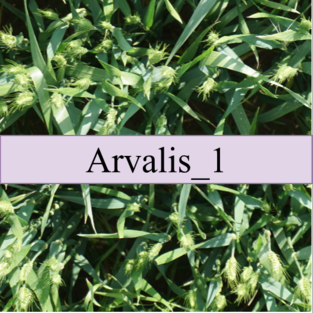}}
			& \BaseModel & YOLOv9e   & 60.1 \\
			& \ROAModel  & \SYNModel & 99.3 \\ \hline
			
			\multirow{2}{*}{\includegraphics[width=\CellWidth, height=\CellHeight]{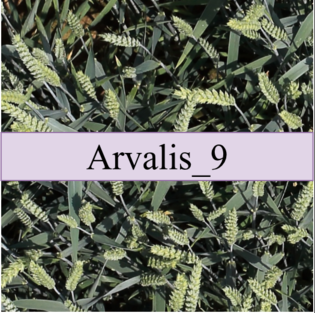}}
			& \BaseModel & YOLOv9e   & 29.9 \\
			& \ROAModel  & \SYNModel & 99.5 \\ \hline
			
			\multirow{2}{*}{\includegraphics[width=\CellWidth, height=\CellHeight]{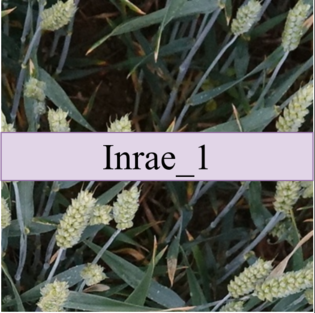}}
			& \BaseModel & YOLOv9e   & 35.5 \\
			& \ROAModel  & \SYNModel & 99.0 \\ \hline
			
			\multirow{2}{*}{\includegraphics[width=\CellWidth, height=\CellHeight]{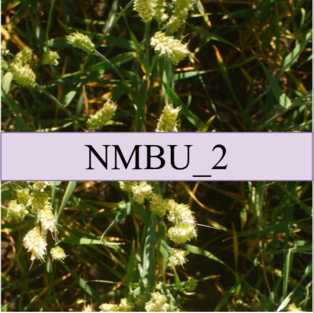}}
			& \BaseModel & YOLOv9e   & 28.1 \\
			& \ROAModel  & \SYNModel & 97.7 \\ \hline
		\end{tabular}
	\end{minipage}%
	\hspace{0.04\textwidth}
	% --- RIGHT COLUMN: Domains 09-17 ---
	\begin{minipage}[b]{0.48\textwidth}
		\centering
		\begin{tabular}{c c c c}
			Domain & Model & Pretrained & mAP@50 \\ \hline\hline
			
			\multirow{2}{*}{\includegraphics[width=\CellWidth, height=\CellHeight]{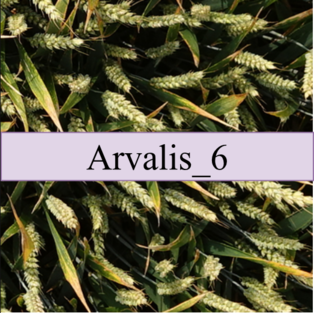}}
			& \BaseModel & YOLOv9e   & 53.4 \\
			& \ROAModel  & \SYNModel & 98.7  \\ \hline
			
			\multirow{2}{*}{\includegraphics[width=\CellWidth, height=\CellHeight]{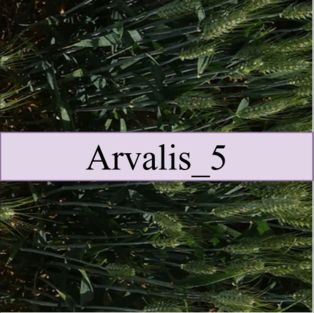}}
			& \BaseModel & YOLOv9e   & 16.4 \\
			& \ROAModel  & \SYNModel & 98.4 \\ \hline
			
			\multirow{2}{*}{\includegraphics[width=\CellWidth, height=\CellHeight]{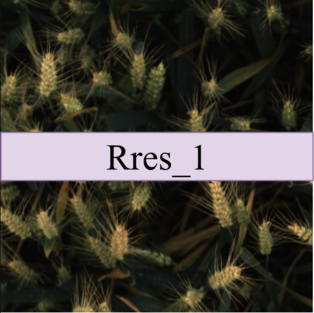}}
			& \BaseModel & YOLOv9e   & 50.7 \\
			& \ROAModel  & \SYNModel & 99.0 \\ \hline
			
			\multirow{2}{*}{\includegraphics[width=\CellWidth, height=\CellHeight]{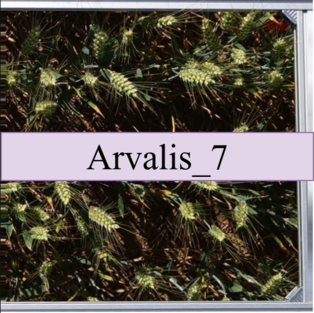}}
			& \BaseModel & YOLOv9e   & 33.2 \\
			& \ROAModel  & \SYNModel & 99.3 \\ \hline
			
			\multirow{2}{*}{\includegraphics[width=\CellWidth, height=\CellHeight]{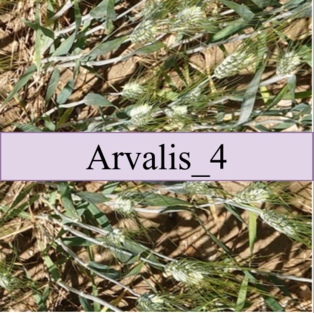}}
			& \BaseModel & YOLOv9e   & 08.7 \\
			& \ROAModel  & \SYNModel & 99.5 \\ \hline
			
			\multirow{2}{*}{\includegraphics[width=\CellWidth, height=\CellHeight]{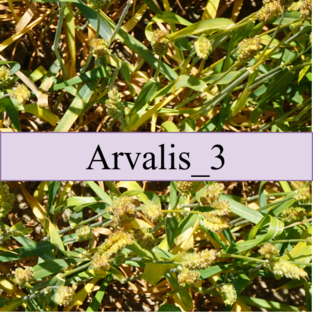}}
			& \BaseModel & YOLOv9e   & 24.9 \\
			& \ROAModel  & \SYNModel & 98.3 \\ \hline
			
			\multirow{2}{*}{\includegraphics[width=\CellWidth, height=\CellHeight]{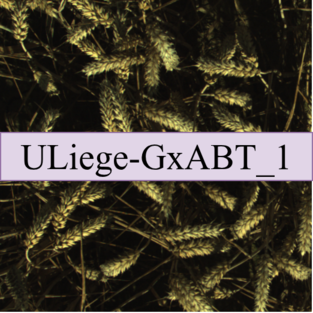}}
			& \BaseModel & YOLOv9e   & 62.6 \\
			& \ROAModel  & \SYNModel & 95.8 \\ \hline
			
			\multirow{2}{*}{\includegraphics[width=\CellWidth, height=\CellHeight]{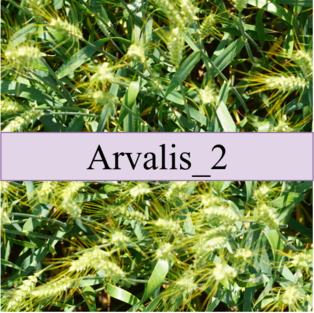}}
			& \BaseModel & YOLOv9e   & 34.4 \\
			& \ROAModel  & \SYNModel & 98.9 \\ \hline
			
			\multirow{2}{*}{\includegraphics[width=\CellWidth, height=\CellHeight]{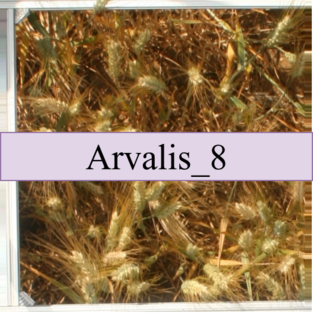}}
			& \BaseModel & YOLOv9e   & 18.1 \\
			& \ROAModel  & \SYNModel & 99.5 \\ \hline
		\end{tabular}
	\end{minipage}
\end{table}

To evaluate the generalization capability and robustness of the proposed framework, we conducted a comprehensive per-domain performance analysis across the $18$ diverse acquisition environments in the~\TestGlobalDomain dataset. As summarized in Table~\ref{tab:PerDomainPerformance}, the \ROAModel demonstrates significant and consistent performance gains over the RGB \BaseModel in every evaluated domain. While the~\BaseModel exhibits high variance and struggles in several challenging environments (\textit{Arvalis\_5} and \textit{Arvalis\_4}), our approach maintains high $mAP@50$ scores, effectively bridging the domain gap through the combination of~\ColorMap representation and rotation-augmented adaptation. \par

\begin{figure}[!th]
    \centering
    \includegraphics[width=\textwidth]{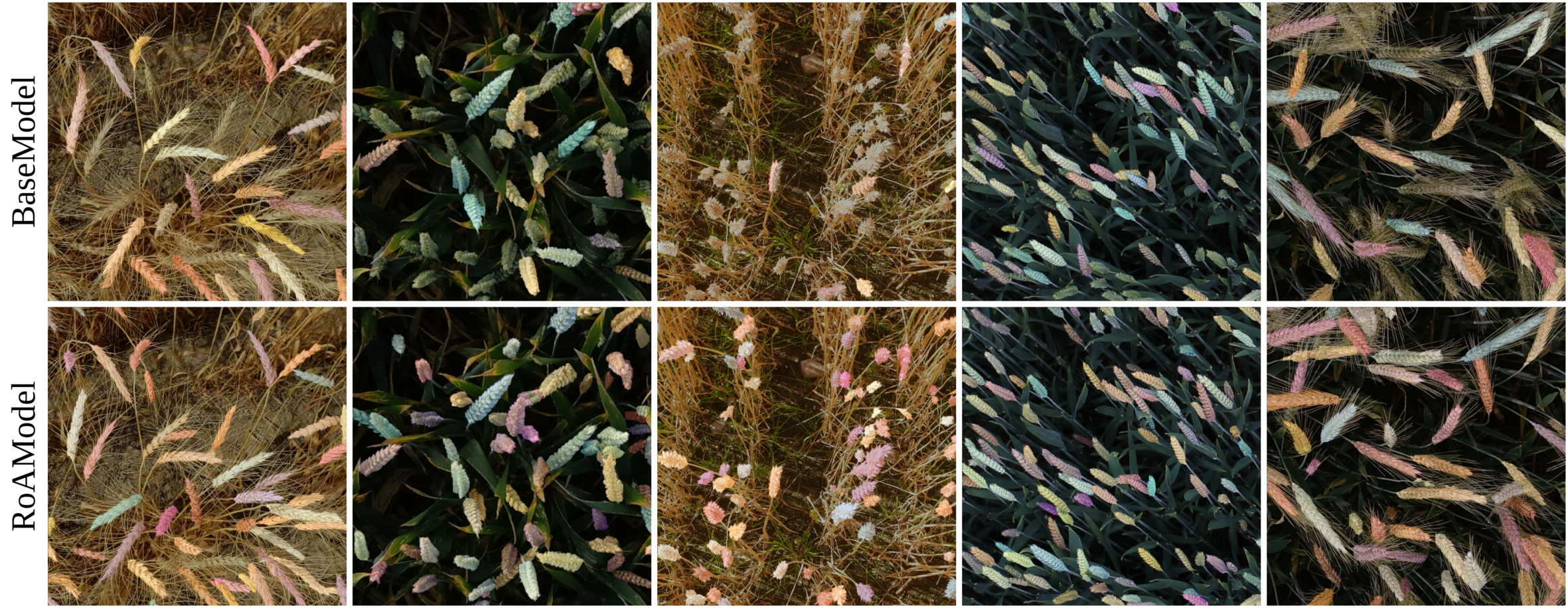}
    \caption{Qualitative performance comparison between the~\BaseModel and our best model~\ROAModel. The first column presents an image from~\TestLateDomain while the rest of the columns display samples chosen from~\TestGlobalDomain. The prediction instance masks are overlaid on the images.
    }
    \label{fig:06_test_vis_per}
\end{figure}

\begin{figure}[!tbph]
    \centering
    \includegraphics[width=\textwidth]{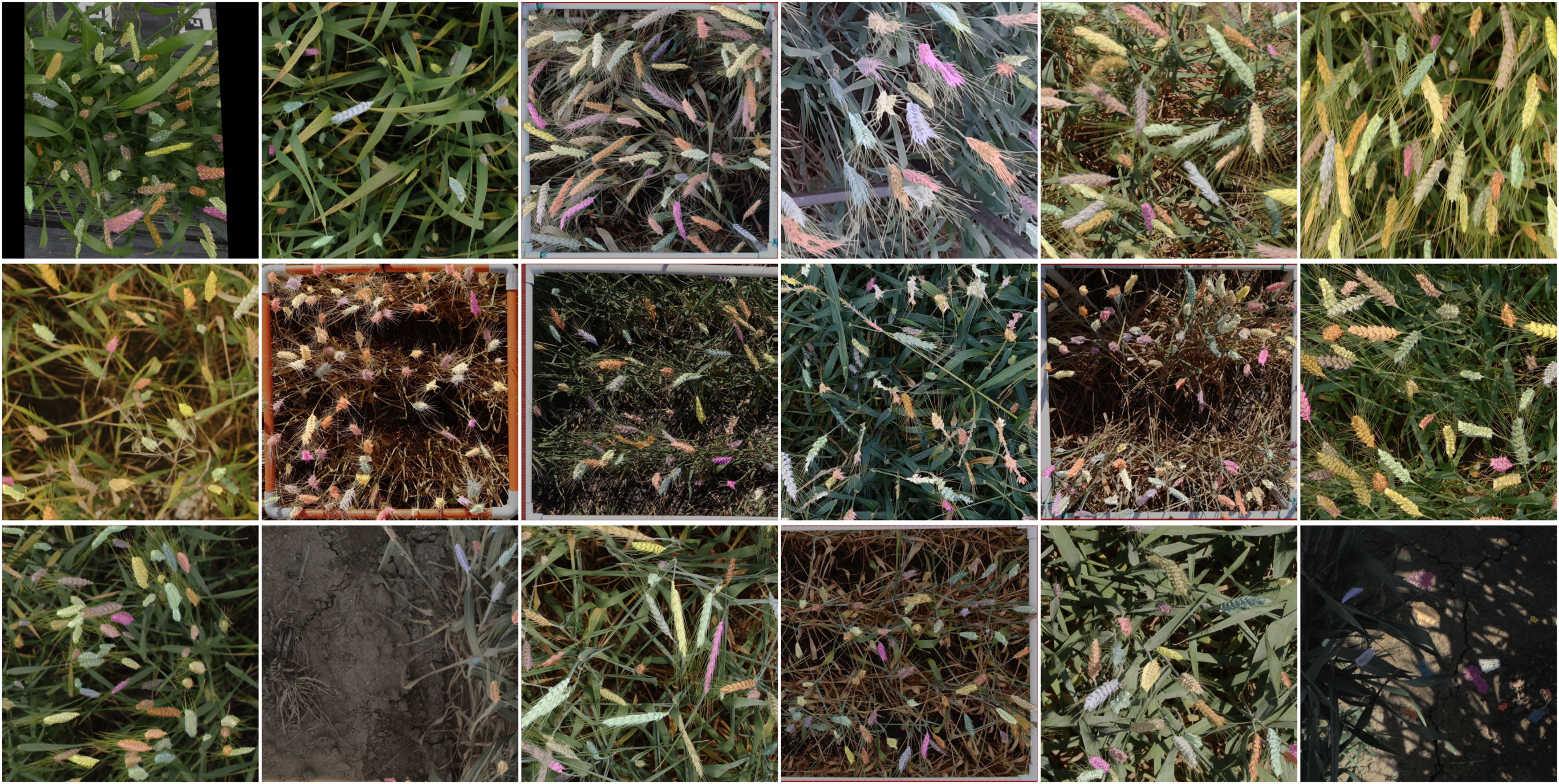}
    \caption{Prediction performance of~\ROAModel on across $18$ domains within one of our external quality assessment set,~\PredictionGlobalTest.
    }
    \label{fig:07_quality_asses_original_gwhd_testset_18_domains}
\end{figure}

\begin{figure}[!th]
    \centering
    \includegraphics[width=\textwidth]{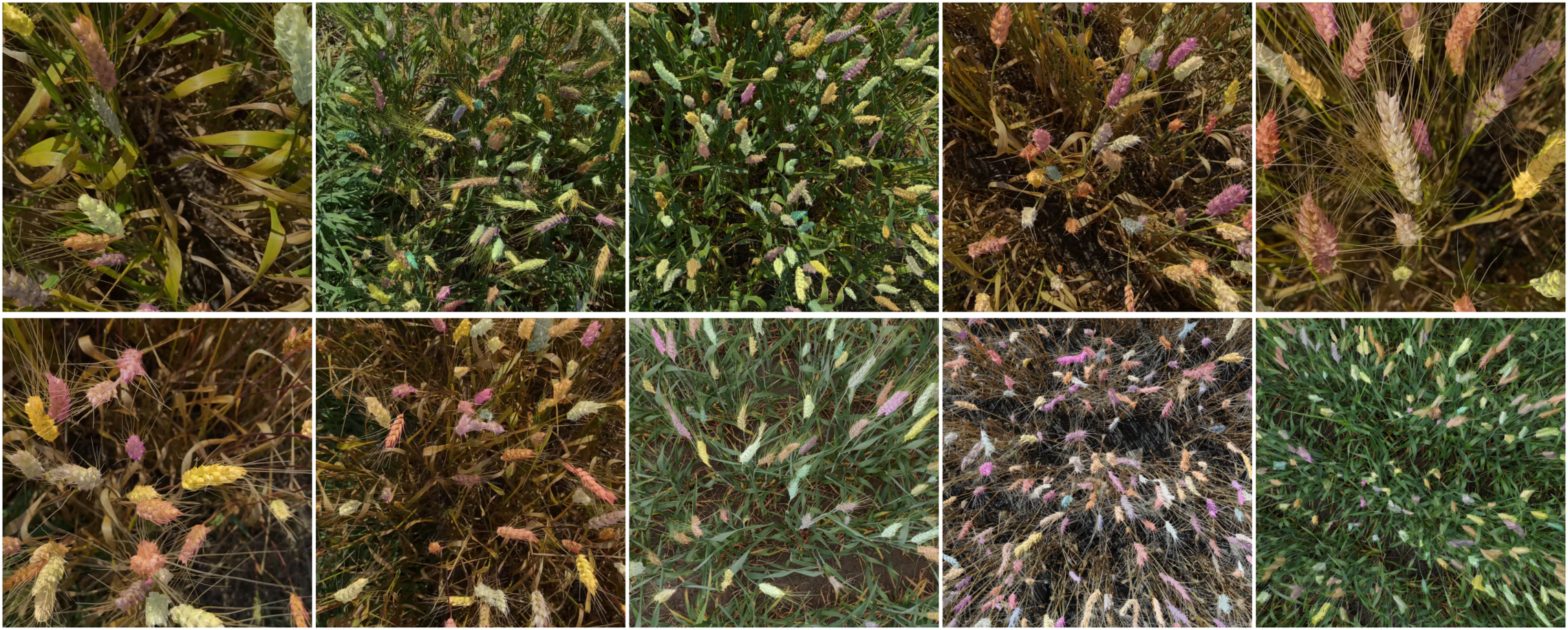}
    \caption{Prediction quality of \ROAModel across the $10$ domains of our external handheld-captured quality assessment set, \PredictionHandHeld.}
    \label{fig:08_more_test_vis_held}
\end{figure}

Considering the better mAP@50-95 scores of this model on both sets, Figure~\ref{fig:06_test_vis_per} visually presents the prediction quality of the~\ROAModel in comparison to~\BaseModel on a few samples chosen from~\TestLateDomain and~\TestGlobalDomain test sets. Regarding the quality assessment on~\PredictionHandHeld and~\PredictionGlobalTest,~\ROAModel visually demonstrates substantial accuracy across all $28$ data domains (please see Figures~\ref{fig:07_quality_asses_original_gwhd_testset_18_domains} and ~\ref{fig:08_more_test_vis_held}). We conducted an additional comparative experiment by training~\PSEModel on~\SYNModel, using~\PseudoSetTrain and~\PseudoSetValid. This model underperformed~\ROAModel when evaluated on both test sets. \par

\begin{table*}[!th]
\centering
\renewcommand{\arraystretch}{1.2}
\caption{
Performance comparison of the same YOLOv9e architectures trained from scratch using RGB (base model) and \ColorMap (our approach) versions of COCO dataset~\cite{Lin2014MicrosoftCC}.
}
\label{tab:cocomodels_performances}
\resizebox{\textwidth}{!}{%
\begin{tabular}{ccccccc}
    \hline
    \textbf{Model} & \textbf{Pretrained Model} & \textbf{Color Space} & \textbf{P}~$\uparrow$ & \textbf{R}~$\uparrow$ & \textbf{mAP@50}~$\uparrow$ & \textbf{mAP@50-95}~$\uparrow$ \\ \hline
    \multirow{2}{*}{\CocoModel} & None  & RGB & 60.0         & 45.5          & 55.4             & 37.6          \\
                                & None  & \ColorMap & \textbf{70.6} & \textbf{58.7} & \textbf{68.0}   & \textbf{55.4} \\ \hline 
    \end{tabular}
}
\end{table*}

\begin{figure}
    \centering
    \includegraphics[width=\textwidth]{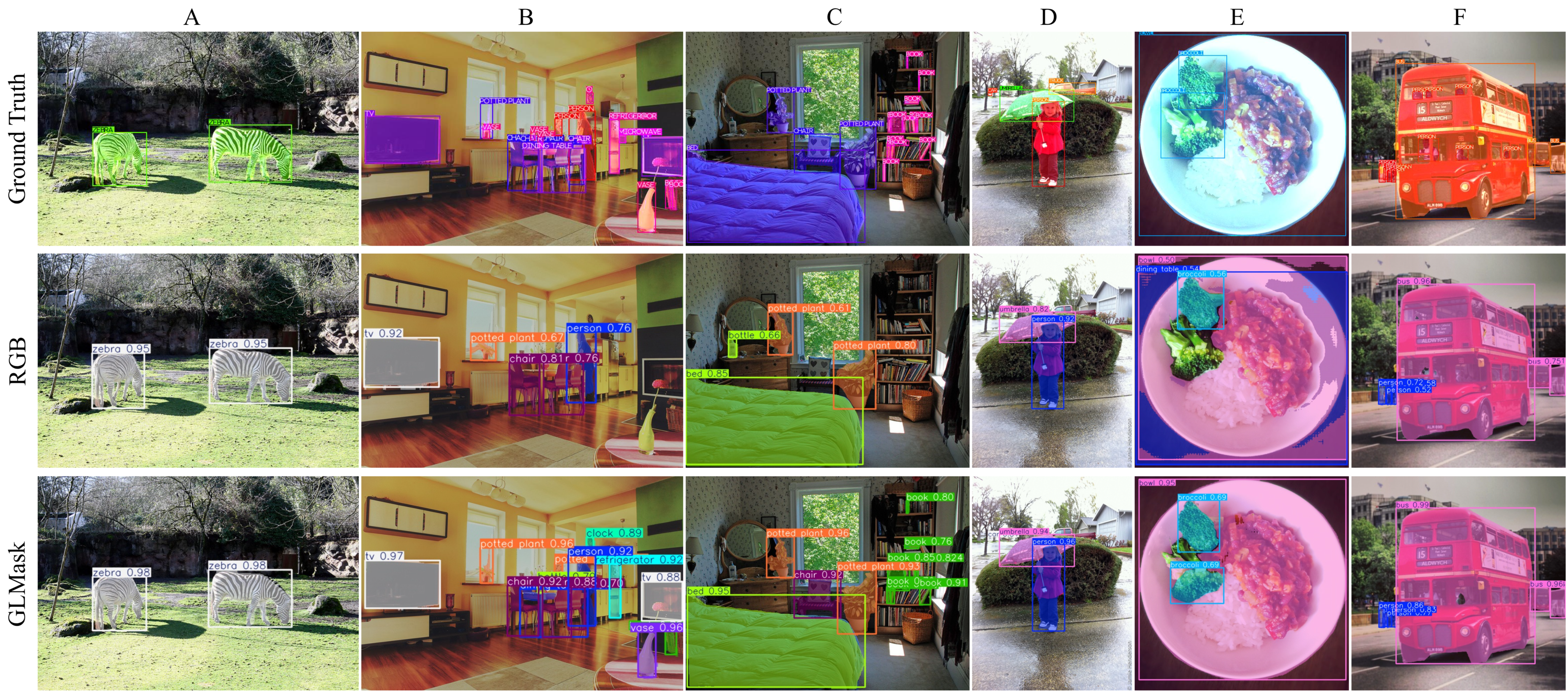}
    \caption{Prediction performance of the COCO models on the Microsoft COCO 2017 dataset~\cite{lin2015microsoft}. Our proposed \ColorMap approach consistently achieved superior segmentation performance (columns B, C, D, and E), and obtained higher confidence scores for detected objects (A through F). In some cases both RGB and \ColorMap models failed to detect objects of interest, examples include the truck in column D and people in F. In addition, near-perfect segmentation performance was observed in cases such as A.
    }
    \label{fig:09_coco_visual_prediction_perfromances}
\end{figure}

We additionally experimented with our proposed approach on a general-purpose dataset and task using the Microsoft COCO 2017 dataset~\cite{Lin2014MicrosoftCC}, extending beyond precision agriculture to assess its applicability across broader domains. We trained two models, one using the dataset's RGB version of training and validation sets and one applying our proposed \ColorMap version. Table~\ref{tab:cocomodels_performances} presents model performances on the validation set of this dataset, illustrating that our proposed approach achieved improvements in up to $12.6\%$ of mAP@50 and over $17.8\%$ in mAP@50-95. 
Additionally, Figure~\ref{fig:09_coco_visual_prediction_perfromances} illustrates a comparison of prediction performances between the RGB and \ColorMap models, displayed in the second and third rows, respectively, with the ground truth shown in the first row. Each column in the figure represents a distinct sample, annotated with the ground truth labels, annotations generated by the base RGB model, and annotations produced by the \ColorMap model developed using our proposed approach.

% -----------------------------------------
\section{Discussion and Limitations}\label{sec:discussion}
In this work, we proposed a semi-self-supervised learning approach for the \InstSeg task of images containing densely overlapped objects, specifically evaluating the methodology in agricultural domains with a focus on wheat head segmentation. Utilizing only $10 + 36$ manually annotated images, we developed a high-performing model that demonstrated strong performance across diverse wheat domains, each characterized by distinct crop phenotypes, canopy structures, and vegetation~\cite{david2021global}. We selected the YOLOv9~\cite{Wang2024YOLOv9LW} model due to its high capacity, speed, and efficiency in image segmentation, making it particularly well-suited for deployment on agricultural machinery performing real-time tasks. It is also important to note that the real-time YOLO architectures are widely recognized as state-of-the-art in instance segmentation, making them a robust and reliable baseline model for our approach. \par 
We illustrated that human \InstSeg annotation of wheat field images and video frames is both challenging and time-consuming. To address this, we generated two large-scale datasets including synthetic, large-scale rotation-augmented, along with their computationally labeled instance masks, to bypass the laborious manual annotation process. Additionally, we introduced~\ColorMap, a novel image representation technique combining grayscale, the L channel of the LAB color map, and computationally or model-generated semantic masks. \ColorMap replaces RGB inputs in the \InstSeg models development, enhancing semantic to instance segmentation transfer while reducing the model's over-reliance on color information. \par 
Using~\ColorMap, \SYNModel, which was trained exclusively on synthetic data, achieved an impressive $97.9$ $mAP@50$ on the external test set of $18$ diverse wheat domains collected around the world, further demonstrating the effectiveness of the proposed data synthesis method in \InstSeg model training. Furthermore, it demonstrates the effectiveness of our proposed approach in enhancing the model's generalizability, allowing it to adapt to various data domains despite being trained on only one domain. To enhance model robustness and generalizability to real data, we implemented a domain adaptation technique, resulting in almost $1\%$ of mAP@50 and over $8\%$ increase in mAP@50-95. 
We further explored pseudo-labeling as an alternative to rotation-augmented methodology. 
However, the results indicated that this approach was less effective in comparison to the rotation-augmented domain adaptation technique. \par 
Beyond the agricultural domain, evaluating our proposed approach on a general-purpose instance segmentation task using the MS COCO dataset~\cite{Lin2014MicrosoftCC} supports its applicability across diverse fields. Although only binary masks, rather than semantic masks, were used to generate the \ColorMap version of the COCO dataset for model development, this approach achieved a $17.8\%$ improvement in performance compared to the baseline YOLO model trained on the RGB version under identical training and evaluation conditions for fair comparison. As our approach originally advocated, the generation of \ColorMap images using semantic masks instead of binary masks could further enhance \CocoModel performance, suggesting that future studies should explore the use of semantic masks and additional color spaces to create \ColorMap images.\par
\begin{figure}[!th]
    \centering
    \includegraphics[width=\textwidth]{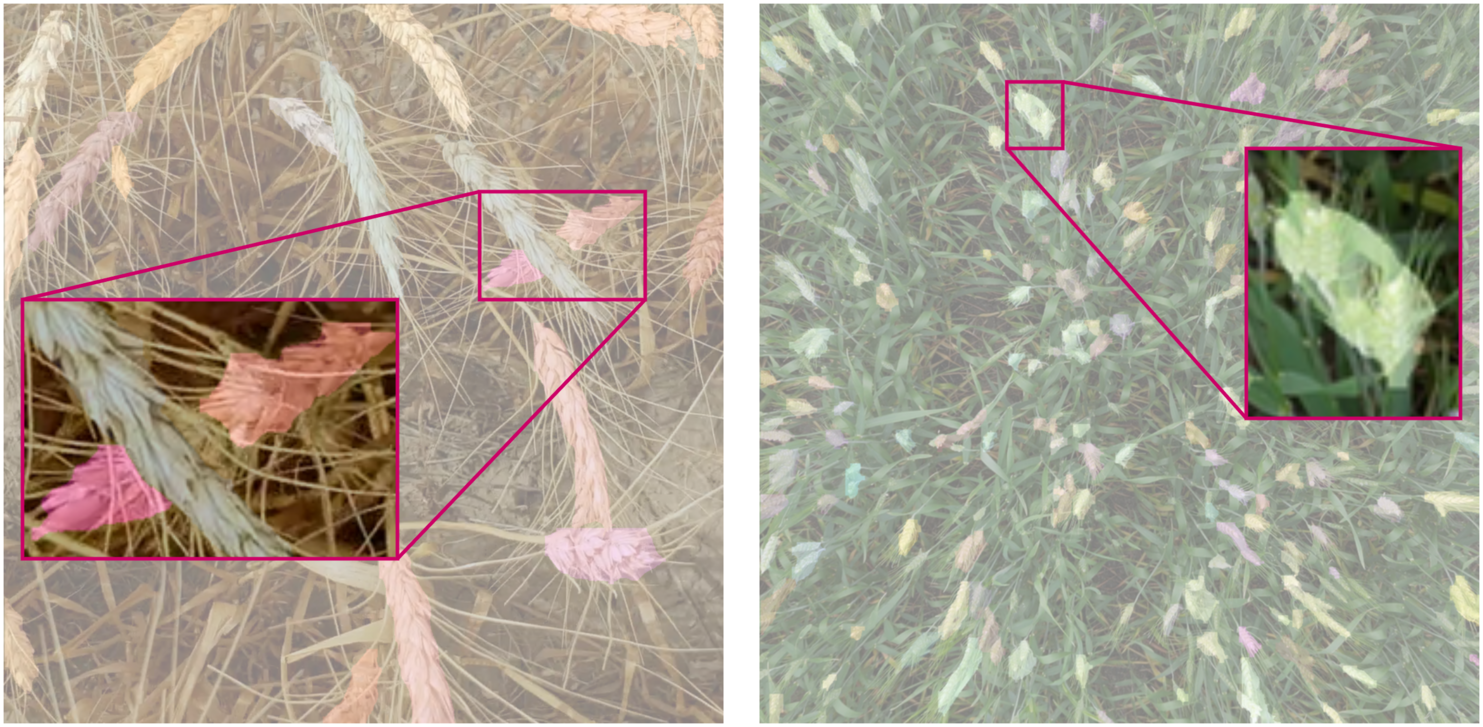}
    \caption{(left) Large objects occluded by other objects resulting in the misidentification of fragmented parts as separate objects. (right) High-altitude images with numerous small wheat heads where overlapping objects are erroneously labeled as a single wheat head.
    }
    \label{fig:10_limitations}
\end{figure}
Despite the advancements, our model exhibits limitations in specific scenarios. For example, when large objects are occluded by another object and become fragmented, the model misidentifies the separated parts as distinct objects. Additionally, in images captured from high altitudes with numerous small wheat heads, the model may merge overlapping objects into a single label, leading to inaccuracies in object identification. These phenomena are visually demonstrated in Figure~\ref{fig:10_limitations}. We suggest further domain-specific model training to address these limitations. \par
In this study, we opted to use the YOLO model, a real-time state-of-the-art model for instance segmentation, to conduct our experiments. This choice aligns with our objective of developing precision agriculture models deployable on agricultural machinery, capable of real-time predictions on larger-scale images, such as $1024 \times 1024$. However, this study lacks an evaluation of our proposed approach on other model architectures, such as SAM variations~\cite{Kirillov2023SegmentA, Ravi2024SAM2S}, which, while resource-intensive for training and inference, may yet have limited applicability across various domains.

% -----------------------------------------
\section{Conclusion}\label{sec:conclusion}
We proposed a semi-self-supervised learning approach for transforming semantic to instance-level segmentation, utilizing a semantic segmentation model to generate semantic masks. We then developed a pipeline that takes an image and its semantic mask as input and produces an instance-level segmentation. We showed the utility of our approach for wheat head instance segmentation; however, our approach could be utilized for similar tasks from other domains as well, as demonstrated through experiments on the COCO dataset. \par

\section*{Acknowledgements}
\begin{small}
\noindent\textbf{Author Contributions:} Conceptualization and Problem Definition, F. Maleki, L. Jin, I. Stavness; Data collection and annotation, K. Najafian; Model development and experimentation, K. Najafian; Visualization, K.N.; Writing original draft preparation and literature review, K. Najafian; Writing—review and editing, F. Maleki, L. Jin, and I. Stavness; resources, L. Jin and I. Stavness; Supervision, F. Maleki, L. Jin, and I. Stavness. All authors have reviewed and approved the final version of the manuscript. \\

\noindent\textbf{Conflict of Interest Statement:} The authors declare that the research was conducted in the absence of any commercial or financial relationships that could be construed as a potential conflict of interest. \\

\noindent\textbf{Funding:} This research is supported by the Natural Sciences and Engineering Research Council of Canada (NSERC), grant number 2019-06424 (LJ), 2019-07017 (IS), and 2024-04966 (FM). \\

\noindent\textbf{Data Statement:} The data and code supporting the findings of this study will be made available upon acceptance of the paper. \\

\noindent\textbf{Sex and Gender Considerations:} This study analyzes agricultural data collected via cameras and drones and does not involve human or animal subjects. As such, sex and gender considerations do not apply to the research. The other dataset used in this study is publicly available and does not require sex or gender-based analysis.\\

\noindent\noindent\textbf{Ethical Considerations:} This study does not involve human or animal subjects. It utilizes agricultural data collected through cameras and a widely recognized, publicly available dataset that does not contain human or animal-related information. Therefore, ethical guidelines concerning human and animal studies are not applicable. \\

\noindent\textbf{Declaration of Generative AI Use:} We declare that no generative AI or AI-assisted technologies were used in the writing of this manuscript. 

\end{small}

%%%%%%%%%%%%%%%%%%%%%%%%%%%%%%%%%%%%%%%%%%
\bibliographystyle{elsarticle-harv} 
\bibliography{references}

@article{ILSVRC15,
Author = {Olga Russakovsky and Jia Deng and Hao Su and Jonathan Krause and Sanjeev Satheesh and Sean Ma and Zhiheng Huang and Andrej Karpathy and Aditya Khosla and Michael Bernstein and Alexander C. Berg and Li Fei-Fei},
Title = {{ImageNet Large Scale Visual Recognition Challenge}},
Year = {2015},
journal   = {International Journal of Computer Vision ({IJCV})},
doi = {10.1007/s11263-015-0816-y},
volume={115},
number={3},
pages={211-252}
}

@article{najafian2023semi,
  title={Semi-Self-Supervised Learning for Semantic Segmentation in Images with Dense Patterns},
  author={Najafian, Keyhan and Ghanbari, Alireza and Sabet Kish, Mahdi and Eramian, Mark and Shirdel, Gholam Hassan and Stavness, Ian and Jin, Lingling and Maleki, Farhad},
  journal={Plant Phenomics},
  volume={5},
  pages={0025},
  year={2023},
  publisher={AAAS}
}

@inproceedings{najafian2021semi,
  title={A Semi-self-supervised Learning Approach for Wheat Head Detection using Extremely Small Number of Labeled Samples},
  author={Najafian, Keyhan and Ghanbari, Alireza and Stavness, Ian and Jin, Lingling and Shirdel, Gholam Hassan and Maleki, Farhad},
  booktitle={Proceedings of the {IEEE/CVF} International Conference on Computer Vision},
  pages={1342--1351},
  year={2021}
}

@article{reis2023real,
  title={Real-Time Flying Object Detection with YOLOv8},
  author={Reis, Dillon and Kupec, Jordan and Hong, Jacqueline and Daoudi, Ahmad},
  journal={arXiv preprint arXiv:2305.09972},
  year={2023}
}

@inproceedings{ronneberger2015u,
  title={U-Net: Convolutional Networks for Biomedical Image Segmentation},
  author={Ronneberger, Olaf and Fischer, Philipp and Brox, Thomas},
  booktitle={Medical Image Computing and Computer-Assisted Intervention--MICCAI 2015: 18th International Conference, Munich, Germany, October 5-9, 2015, Proceedings, Part III 18},
  pages={234--241},
  year={2015},
  organization={Springer}
}

@article{fourati2021wheat,
  title={Wheat head detection using deep, semi-supervised and ensemble learning},
  author={Fourati, Fares and Mseddi, Wided Souidene and Attia, Rabah},
  journal={Canadian Journal of Remote Sensing},
  volume={47},
  number={2},
  pages={198--208},
  year={2021},
  publisher={Taylor \& Francis}
}

@article{david2021global,
  title={Global Wheat Head Detection 2021: An Improved Dataset for Benchmarking Wheat Head Detection Methods},
  author={David, Etienne and Serouart, Mario and Smith, Daniel and Madec, Simon and Velumani, Kaaviya and Liu, Shouyang and Wang, Xu and Pinto, Francisco and Shafiee, Shahameh and Tahir, Izzat SA and others},
  journal={Plant Phenomics},
  volume={2021},
  year={2021},
  publisher={{AAAS}}
}

@inproceedings{he2017mask,
  title={Mask R-CNN},
  author={He, Kaiming and Gkioxari, Georgia and Doll{\'a}r, Piotr and Girshick, Ross},
  booktitle={Proceedings of the {IEEE} International Conference on Computer Vision},
  pages={2961--2969},
  year={2017}
}

@inproceedings{chen2021understanding,
  title={Understanding and Mitigating Annotation Bias in Facial Expression Recognition},
  author={Chen, Yunliang and Joo, Jungseock},
  booktitle={Proceedings of the {IEEE/CVF} International Conference on Computer Vision},
  pages={14980--14991},
  year={2021}
}

@inproceedings{hendrycks2021many,
  title={The Many Faces of Robustness: A Critical Analysis of Out-of-Distribution Generalization},
  author={Hendrycks, Dan and Basart, Steven and Mu, Norman and Kadavath, Saurav and Wang, Frank and Dorundo, Evan and Desai, Rahul and Zhu, Tyler and Parajuli, Samyak and Guo, Mike and others},
  booktitle={Proceedings of the {IEEE/CVF} International Conference on Computer Vision},
  pages={8340--8349},
  year={2021}
}

@article{Wang2021GeneralizingTU,
  title={Generalizing to Unseen Domains: A Survey on Domain Generalization},
  author={Jindong Wang and Cuiling Lan and Chang Liu and Yidong Ouyang and Tao Qin},
  journal={IEEE Transactions on Knowledge and Data Engineering},
  year={2021},
  volume={35},
  pages={8052-8072},
  url={https://api.semanticscholar.org/CorpusID:232110832}
}

@article{Wang2024YOLOv9LW,
  title={YOLOv9: Learning What You Want to Learn Using Programmable Gradient Information},
  author={Chien-Yao Wang and I-Hau Yeh and Hongpeng Liao},
  journal={ArXiv},
  year={2024},
  volume={abs/2402.13616},
  url={https://api.semanticscholar.org/CorpusID:267770251}
}

@article{Wang2024InstanceSF,
  title={Instance Segmentation Frustum–PointPillars: A Lightweight Fusion Algorithm for Camera–LiDAR Perception in Autonomous Driving},
  author={Yongsheng Wang and Xiaobo Han and Xiaoxu Wei and Jie Luo},
  journal={Mathematics},
  year={2024},
  url={https://api.semanticscholar.org/CorpusID:266850655}
}

@article{Bolya2019YOLACTRI,
  title={YOLACT: Real-Time Instance Segmentation},
  author={Daniel Bolya and Chong Zhou and Fanyi Xiao and Yong Jae Lee},
  journal={2019 IEEE/CVF International Conference on Computer Vision (ICCV)},
  year={2019},
  pages={9156-9165},
  url={https://api.semanticscholar.org/CorpusID:102354230}
}

@article{Wang2022YOLOv7TB,
  title={YOLOv7: Trainable Bag-of-Freebies Sets New State-of-the-Art for Real-Time Object Detectors},
  author={Chien-Yao Wang and Alexey Bochkovskiy and Hong-Yuan Mark Liao},
  journal={2023 IEEE/CVF Conference on Computer Vision and Pattern Recognition (CVPR)},
  year={2022},
  pages={7464-7475},
  url={https://api.semanticscholar.org/CorpusID:250311206}
}

@article{Zhao2023FastSA,
  title={Fast Segment Anything},
  author={Xu Zhao and Wen-Yan Ding and Yongqi An and Yinglong Du and Tao Yu and Min Li and Ming Tang and Jinqiao Wang},
  journal={ArXiv},
  year={2023},
  volume={abs/2306.12156},
  url={https://api.semanticscholar.org/CorpusID:259212104}
}

@article{Kirillov2023SegmentA,
  title={Segment Anything},
  author={Alexander Kirillov and Eric Mintun and Nikhila Ravi and Hanzi Mao and Chloe Rolland and Laura Gustafson and Tete Xiao and Spencer Whitehead and Alexander C. Berg and Wan-Yen Lo and Piotr Doll{\'a}r and Ross B. Girshick},
  journal={2023 IEEE/CVF International Conference on Computer Vision (ICCV)},
  year={2023},
  pages={3992-4003},
  url={https://api.semanticscholar.org/CorpusID:257952310}
}

@article{Ennadifi2022LocalUW,
  title={Local Unsupervised Wheat Head Segmentation},
  author={Elias Ennadifi and S{\'e}bastien Dandrifosse and Mohammed El Amine Mokhtari and Alexis Carlier and Sohaib Laraba and Beno{\^i}t Mercatoris and Bernard Gosselin},
  journal={2022 IEEE 18th International Conference on Intelligent Computer Communication and Processing (ICCP)},
  year={2022},
  pages={55-62},
  url={https://api.semanticscholar.org/CorpusID:257261366}
}

@article{Nayak2024ImprovedDO,
  title={Improved Detection of Fusarium Head Blight in Wheat Ears through YOLACT Instance Segmentation},
  author={Naman Nayak and Deepak Kumar and Saumitra Chattopadhay and Vinay Kukreja and Aditya Verma},
  journal={2024 11th International Conference on Reliability, Infocom Technologies and Optimization (Trends and Future Directions) (ICRITO)},
  year={2024},
  pages={1-4},
  url={https://api.semanticscholar.org/CorpusID:269777088}
}

@article{Gao2022AutomaticTD,
  title={Automatic Tandem Dual BlendMask Networks for Severity Assessment of Wheat Fusarium Head Blight},
  author={Yichao Gao and Hetong Wang and Man Li and Wen-Hao Su},
  journal={Agriculture},
  year={2022},
  url={https://api.semanticscholar.org/CorpusID:252401695}
}

@article{Li2023EnhancingAI,
  title={Enhancing Agricultural Image Segmentation with an Agricultural Segment Anything Model Adapter},
  author={Yaqin Li and Dandan Wang and Cao Yuan and Hao Li and Jing Hu},
  journal={Sensors (Basel, Switzerland)},
  year={2023},
  volume={23},
  url={https://api.semanticscholar.org/CorpusID:261988826}
}

@article{thisanke2023semantic,
  title={Semantic segmentation using Vision Transformers: A survey},
  author={Thisanke, Hans and Deshan, Chamli and Chamith, Kavindu and Seneviratne, Sachith and Vidanaarachchi, Rajith and Herath, Damayanthi},
  journal={Engineering Applications of Artificial Intelligence},
  volume={126},
  pages={106669},
  year={2023},
  publisher={Elsevier}
}

@article{Deng2023SegmentAM,
  title={Segment Anything Model (SAM) for Digital Pathology: Assess Zero-shot Segmentation on Whole Slide Imaging},
  author={Ruining Deng and Can Cui and Quan Liu and Tianyuan Yao and Lucas W. Remedios and Shunxing Bao and Bennett A. Landman and Lee E. Wheless and Lori A. Coburn and Keith T. Wilson and Yaohong Wang and Shilin Zhao and Agnes B. Fogo and Haichun Yang and Yucheng Tang and Yuankai Huo},
  journal={ArXiv},
  year={2023},
  volume={abs/2304.04155},
  url={https://api.semanticscholar.org/CorpusID:258049163}
}

@inproceedings{Lin2014MicrosoftCC,
  title={Microsoft COCO: Common Objects in Context},
  author={Tsung-Yi Lin and Michael Maire and Serge J. Belongie and James Hays and Pietro Perona and Deva Ramanan and Piotr Doll{\'a}r and C. Lawrence Zitnick},
  booktitle={European Conference on Computer Vision},
  year={2014},
  url={https://api.semanticscholar.org/CorpusID:14113767}
}

@article{CossioMontefinale2024CherryCD,
  title={Cherry CO Dataset: A Dataset for Cherry Detection, Segmentation and Maturity Recognition},
  author={Luis Cossio-Montefinale and Javier Ruiz-del-Solar and Rodrigo Verschae},
  journal={{IEEE} Robotics and Automation Letters},
  year={2024},
  volume={9},
  pages={5552-5558},
  url={https://api.semanticscholar.org/CorpusID:269400278}
}

@article{Marinello2023ThePT,
  title={The Path to Smart Farming: Innovations and Opportunities in Precision Agriculture},
  author={Francesco Marinello and Xiuguo Zou and Zheng Liu and Xiaochen Zhu and Wentian Zhang and Yan Qian and Yuhua Li and E.M.B.M. Karunathilake and Anh Tuan Le and Seong Heo and Yong Suk Chung and Sheikh Mansoor},
  journal={Agriculture},
  year={2023},
  url={https://api.semanticscholar.org/CorpusID:260852984}
}

@article{Chin2023PlantDD,
  title={Plant disease detection using drones in precision agriculture},
  author={Ruben Chin and Cagatay Catal and Ayalew Kassahun},
  journal={Precision Agriculture},
  year={2023},
  volume={24},
  pages={1663-1682},
  url={https://api.semanticscholar.org/CorpusID:257815513}
}

@article{Thangaraj2023ACS,
  title={A comparative study of deep learning and Internet of Things for precision agriculture},
  author={Saranya Thangaraj and Deisy Chelliah and Sridevi Subbiah and Kalaiarasi Sonai Muthu Anbananthen},
  journal={Engineering Applications of Artificial Intelligence},
  year={2023},
  volume={122},
  pages={106034},
  url={https://api.semanticscholar.org/CorpusID:257381420}
}

@article{Charisis2024DeepLI,
  title={Deep learning-based instance segmentation architectures in agriculture: A review of the scopes and challenges},
  author={Christos Charisis and Dimitrios Argyropoulos},
  journal={Smart Agricultural Technology},
  year={2024},
  url={https://api.semanticscholar.org/CorpusID:269144172}
}

@article{Sapkota2024ComparingYA,
  title={Comparing YOLOv8 and Mask R-CNN for instance segmentation in complex orchard environments},
  author={Ranjan Sapkota and Dawood Ahmed and Manoj Karkee},
  journal={Artificial Intelligence in Agriculture},
  year={2024},
  url={https://api.semanticscholar.org/CorpusID:271278867}
}

@article{Yue2023ImprovedYN,
  title={Improved YOLOv8-Seg Network for Instance Segmentation of Healthy and Diseased Tomato Plants in the Growth Stage},
  author={Xiang Yue and Kai Qi and Xinyi Na and Yang Zhang and Yanhua Liu and Cuihong Liu},
  journal={Agriculture},
  year={2023},
  url={https://api.semanticscholar.org/CorpusID:261111698}
}

@article{Vyas2023AdvancingPA,
  title={Advancing Precision Agriculture: Leveraging YOLOv8 for Robust Deep Learning Enabled Crop Diseases Detection},
  author={Narayan Vyas and Vishal Dutt},
  journal={2023 International Conference on Integration of Computational Intelligent System ({ICICIS})},
  year={2023},
  pages={1-6},
  url={https://api.semanticscholar.org/CorpusID:267771763}
}

@article{Ghanbari2024SemiSelfSupervisedDA,
  title={Semi-Self-Supervised Domain Adaptation: Developing Deep Learning Models with Limited Annotated Data for Wheat Head Segmentation},
  author={Alireza Ghanbari and Gholam Hassan Shirdel and Farhad Maleki},
  journal={ArXiv},
  year={2024},
  volume={abs/2405.07157},
  url={https://api.semanticscholar.org/CorpusID:269757575}
}

@article{Myers2024EfficientWH,
  title={Efficient Wheat Head Segmentation with Minimal Annotation: A Generative Approach},
  author={Jaden Myers and Keyhan Najafian and Farhad Maleki and Katie L. Ovens},
  journal={Journal of Imaging},
  year={2024},
  volume={10},
  url={https://api.semanticscholar.org/CorpusID:270715030}
}

@article{Gupta2023MulticlassWI,
  title={Multiclass weed identification using semantic segmentation: An automated approach for precision agriculture},
  author={Sanjay Kumar Gupta and Shivam Kumar Yadav and Sanjay Kumar Soni and Udai Shanker and Pradeep Kumar Singh},
  journal={Ecol. Informatics},
  year={2023},
  volume={78},
  pages={102366},
  url={https://api.semanticscholar.org/CorpusID:265197848}
}

@article{Marks2023HighPL,
  title={High Precision Leaf Instance Segmentation for Phenotyping in Point Clouds Obtained Under Real Field Conditions},
  author={Elias Marks and Matteo Sodano and Federico Magistri and Louis Wiesmann and Dhagash Desai and Rodrigo Marcuzzi and Jens Behley and C. Stachniss},
  journal={{IEEE} Robotics and Automation Letters},
  year={2023},
  volume={8},
  pages={4791-4798},
  url={https://api.semanticscholar.org/CorpusID:259325393}
}

@article{Rawat2022HowUI,
  title={How Useful Is Image-Based Active Learning for Plant Organ Segmentation?},
  author={Shivangana Rawat and Akshay L Chandra and Sai Vikas Desai and Nandhini Vineeth and Balasubramanian and Seishi Ninomiya and Wei Guo},
  journal={Plant Phenomics},
  year={2022},
  volume={2022},
  url={https://api.semanticscholar.org/CorpusID:246483813}
}

@inproceedings{Loshchilov2017DecoupledWD,
  title={Decoupled Weight Decay Regularization},
  author={Ilya Loshchilov and Frank Hutter},
  booktitle={International Conference on Learning Representations},
  year={2017},
  url={https://api.semanticscholar.org/CorpusID:53592270}
}

@Article{AlbumentationBuslaev,
    AUTHOR = {Buslaev, Alexander and Iglovikov, Vladimir I. and Khvedchenya, Eugene and Parinov, Alex and Druzhinin, Mikhail and Kalinin, Alexandr A.},
    TITLE = {Albumentations: Fast and Flexible Image Augmentations},
    JOURNAL = {Information},
    VOLUME = {11},
    YEAR = {2020},
    NUMBER = {2},
    ARTICLE-NUMBER = {125},
    URL = {https://www.mdpi.com/2078-2489/11/2/125},
    ISSN = {2078-2489},
    DOI = {10.3390/info11020125}
}

@inproceedings{Poynton1997FrequentlyAQ,
  title={Frequently Asked Questions about Color},
  author={Charles A. Poynton},
  year={1997},
  url={https://api.semanticscholar.org/CorpusID:426113}
}

@inproceedings{Ravi2024SAM2S,
  title={SAM 2: Segment Anything in Images and Videos},
  author={Nikhila Ravi and Valentin Gabeur and Yuan-Ting Hu and Ronghang Hu and Chaitanya K. Ryali and Tengyu Ma and Haitham Khedr and Roman Radle and Chloe Rolland and Laura Gustafson and Eric Mintun and Junting Pan and Kalyan Vasudev Alwala and Nicolas Carion and Chao-Yuan Wu and Ross Girshick and Piotr Doll'ar and Christoph Feichtenhofer},
  year={2024},
  url={https://api.semanticscholar.org/CorpusID:271601113}
}

@misc{lin2015microsoft,
      title={Microsoft COCO: Common Objects in Context},
      author={Tsung-Yi Lin and Michael Maire and Serge Belongie and Lubomir Bourdev and Ross Girshick and James Hays and Pietro Perona and Deva Ramanan and C. Lawrence Zitnick and Piotr Dollár},
      year={2015},
      eprint={1405.0312},
      archivePrefix={arXiv},
      primaryClass={cs.CV}
}

@article{miller1992growth,
  title={Growth stages of wheat},
  author={Miller, Travis D},
  journal={Better crops with plant food. Potash \& Phosphate Institute},
  volume={76},
  pages={12},
  year={1992}
}

\end{document}